\documentclass[10pt,twocolumn,letterpaper]{article}

\usepackage{cvpr}
\usepackage{times}
\usepackage{epsfig}
\usepackage{graphicx}
\usepackage{amsmath}
\usepackage{amssymb}
\usepackage{multirow}
\usepackage{array}
\usepackage{booktabs}
\usepackage{tabularx}
\usepackage{algorithm}
\usepackage{algorithmic}
\usepackage{stmaryrd}
\usepackage[pagebackref=true,breaklinks=true,letterpaper=true,colorlinks,
            allcolors=blue, bookmarks=false]{hyperref}

\cvprfinalcopy 

\def\cvprPaperID{2205}

\ifcvprfinal\fi
\begin{document}

\title{CP-mtML: Coupled Projection multi-task Metric Learning \\ for Large Scale Face Retrieval}

\author{Binod Bhattarai$^{1,}$\thanks{GREYC CNRS UMR 6072. Supported by projects ANR--12--CORD--014--SECULAR  ANR-­‐12-­‐SECU-­‐005--PHYSIONOMIE} \hspace{2em}
Gaurav Sharma$^{2,3,}$\thanks{Currently with CSE, Indian Institute of Technology Kanpur.  Majority of the work was done at
Max Planck Institute for Informatics.} \hspace{2em} Frederic
Jurie$^{1,}$\footnotemark[1] \vspace{0.5em}\\
$^1$University of Caen, France  \hspace{2em} $^2$MPI for Informatics, Germany  
\hspace{2em} $^3$IIT Kanpur, India
}

\maketitle

\def\etal{et al\onedot}
\def\etc{etc\onedot}
\def\ie{i.e\onedot}
\def\eg{e.g\onedot}
\def\cf{cf\onedot}
\def\vs{vs\onedot}
\def\grad{\nabla}
\def\b{\textbf{b}}
\def\v{\textbf{v}}
\def\a{\boldsymbol{\alpha}}
\def\sign{\textrm{sign}}
\def\pd{\partial}
\def\T{\mathcal{T}}
\def\R{\mathbb{R}}
\def\Reg{\mathcal{R}}
\def\X{\mathcal{X}}
\def\I{\mathcal{I}}
\def\F{\mathcal{F}}
\def\Obj{\mathcal{O}}
\def\V{\mathcal{V}}
\def\Lz{L_0}
\def\Lt{L_t}
\def\Dz{{D_0}}
\def\Do{{D_1}}
\def\Dt{{D_2}}
\def\w{\textbf{w}}
\def\c{\textbf{c}}
\def\1{\textbf{1}}
\def\x{\textbf{x}}
\def\c{\textbf{c}}
\def\s{\textbf{s}}
\def\d{\boldsymbol{\delta}}
\def\y{\textbf{y}}
\def\l{\textbf{l}}
\def\ock{$1$-call@$K$}
\def\oc{$1$-call@}
\def\TODO{\textcolor{red}{TODO} }

\begin{abstract}
We propose a novel Coupled Projection multi-task Metric Learning (CP-mtML) method for large scale
face retrieval. In contrast to previous works which were limited to low dimensional features and
small datasets, the proposed method scales to large datasets with high dimensional face descriptors.
It utilises pairwise (dis-)similarity constraints as supervision and hence does not require
exhaustive class annotation for every training image. While, traditionally, multi-task learning
methods have been validated on same dataset but different tasks, we work on the more challenging
setting with heterogeneous datasets and different tasks. We show empirical validation on multiple
face image datasets of different facial traits, \eg identity, age and expression. We use classic
Local Binary Pattern (LBP) descriptors along with the recent Deep Convolutional Neural Network (CNN)
features. The experiments clearly demonstrate the scalability and improved performance of the
proposed method on the tasks of identity and age based face image retrieval compared to competitive
existing methods, on the standard datasets and with the presence of a million distractor face
images.
\end{abstract}

\section{Introduction}
\label{sec:intro}

Many computer vision algorithms heavily rely on a distance function over image signatures and their
performance strongly depends on the quality of the metric.  Metric learning (ML) \ie learning an
optimal distance function for a given task, using annotated training data, is in such cases, a key
to good performance. Hence, ML  has been a very active topic of interest in the machine learning
community and has been widely used in many computer vision algorithms for image annotation
\cite{GuillauminCVPR2009}, person re-identification \cite{BedagkarIVC2014} or face matching
\cite{GuillauminICCV2009}, to mention a few of them.

\begin{figure}[t]
\centering
\includegraphics[width=\columnwidth, trim=0 15 0 -5, clip]{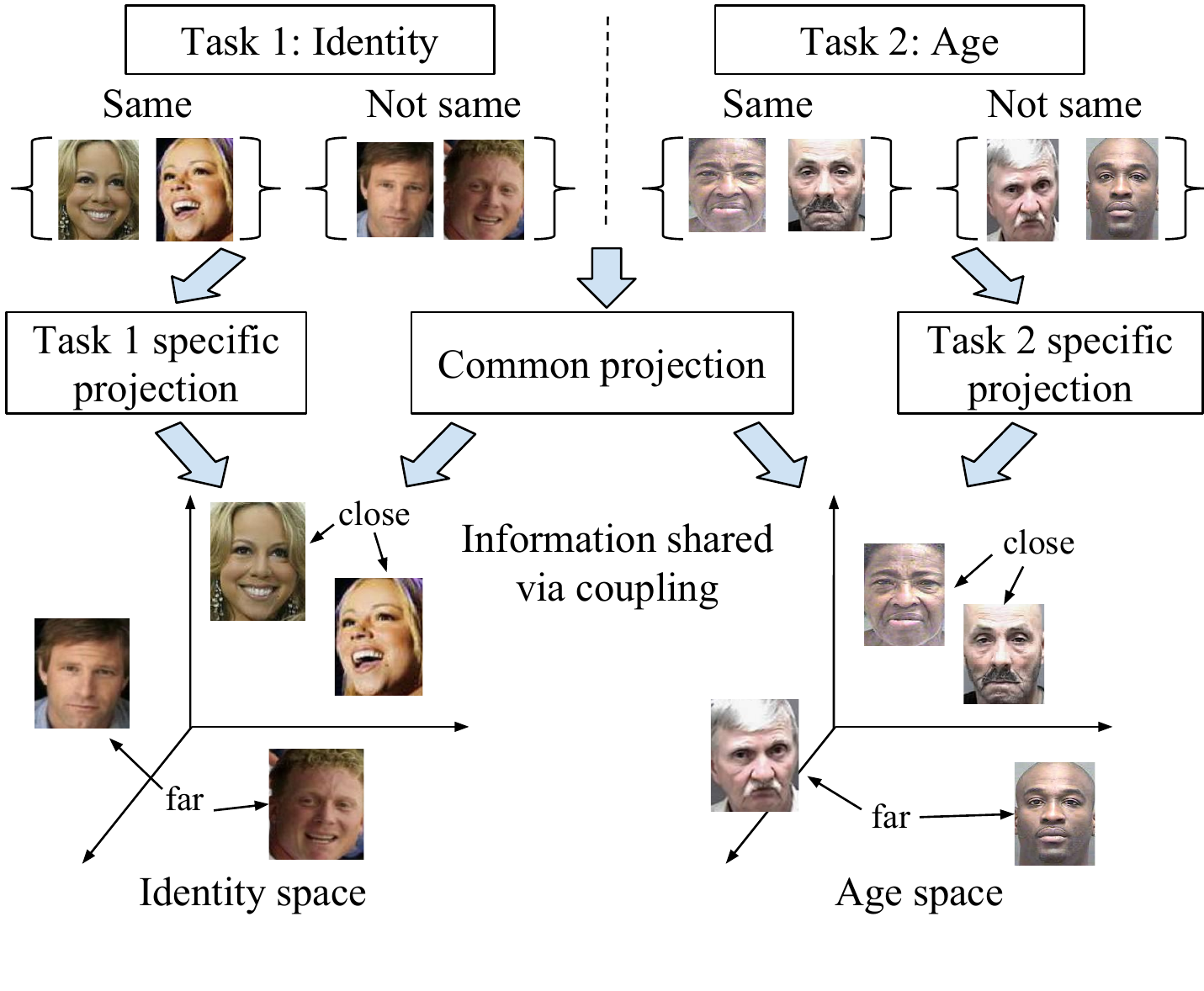} 
\caption{
Illustration of the proposed method. We propose a multi-task metric learning method which learns a
distance function as a projection into a low dimensional Euclidean space, from pairwise
(dis-)similarity constraints. It learns two types of projections jointly: (i) a common projection
shared by all the tasks and (ii) task related specific projections. The final projection for each
task is given by a combination of the common projection and the task specific projection. By
coupling the projections and learning them jointly, the information shared between the related tasks
can lead to improved performance. 
}
\vspace{-1.2em}
\label{fig:illus}
\end{figure}

This paper focuses on the task of face matching \ie comparing images of two faces with respect to
different criteria such as identity, expression or age. More precisely, the task is to retrieve
faces similar to a query, according to the given criteria (\eg identity) and rank them using their
distances to the query. 

One key contribution of this paper is the introduction of a cross-dataset multi-task ML approach.
The main advantage of multi-task ML is leveraging the performance of single task ML by combining
data coming from different but related tasks. While many recent works on classification have shown
that learning metrics for related tasks together using multi-task learning approaches can lead to
improvements in performance \cite{argyriou2008convex,caruana1997multitask, lapin2014scalable,
maurer2013sparse,romera2012exploiting,zhang2014facial},  most of earlier works on face matching are
based on a single task. In addition, there are only a few works on multi-task ML
\cite{parameswaran2010large, wang2012multi, yang2013multi}, with most of the multi-task approaches being
focussed on multi-task classification. In addition, the previous multi-task ML methods have been
shown to work on the same dataset but not on cross dataset problems. Finally, none of the mentioned
approaches have been showed to be scalable to millions of images with features of thousands of
dimensions.  

In the present paper, our goal is hence to develop a scalable multi-task ML method, using linear
embeddings for dimensionality reduction, able to leverage related tasks from heterogeneous
datasets/sources of faces. Such challenging multi-task heterogeneous dataset setting, while being a
very practical setting, has received almost negligible attention in the literature. Towards that
goal, this paper presents a novel Coupled Projection multi-task Metric Learning method (CP-mtML) for
learning better distance metrics for a main task by utilizing additional data from related
auxiliary tasks. The method works with pairwise supervision of similar and dissimilar faces -- in
terms of different aspects \eg identity, age and expression -- and does not require exhaustive
annotation with presence or absence of classes for all images. We pose the metric learning task as
the one of learning coupled low dimensional projections, one for each task, where the final distance
is given by the Euclidean distance in the respective projection spaces. 
 
The projections are coupled with each other by enforcing them to be a combination of a common
projection and a task specific one. The common projection is expected to capture the commonalities
in the different tasks, while the task specific components are expected to specialize to the
specificities of the corresponding tasks. The projections are jointly learned using, at the same
time, training data from different datasets containing different tasks. 

The proposed approach is experimentally validated with challenging publicly available datasets for
facial analysis based on identity, age and expression. The task of semantic face retrieval is
evaluated in a large scale setting, \ie in the presence of order of millions of distractors, and
compared with challenging baselines based on state-of-the-art unsupervised and supervised projection
learning methods. The proposed model consistently improves over the baselines. The experimental
section also provides qualitative results visually demonstrating the improvement of the method over
the most challenging baselines.

\section{Related Work}
%
%

As said in the introduction, because of its key role in many problems, ML has received lot of
attention in the literature. The reader can refer to  \cite{BelletArxiv2013, KulisFTML2012} for
comprehensive surveys on ML approaches in general.   Among the possible classes of distances,
the Mahalanobis-like one is certainly the most widely studied \cite{MignonCVPR2012,
SalakhutdinovAISTATS2007, weinberger2009distance, XingNIPS2002} and has been very successful in
variety of face matching tasks \cite{bhattarai2014some, GuillauminCVPR2009,
GuillauminICCV2009, Simonyan13}. 

The various Mahalanobis-like methods differ in their objective functions which are themselves related to the type of constraints provided by the training data.  The constraints can be given at class level (\ie same-class vectors have to be close from one another after projection) 
\cite{SalakhutdinovAISTATS2007}, under the form of triplet constraints \ie $(\x_i,\x_j,\x_k)$ with $\x_i$ relatively
closer to $\x_j$ compared to $\x_k$ \cite{weinberger2009distance}, or finally by pairwise constraints
$(\x_i,\x_j,y_{ij})$ such that $\x_i$ and $\x_j$ are similar (dissimilar) if $y_{ij}=+1$
($y_{ij}=-1)$ \cite{MignonCVPR2012,Simonyan13}.

While the above mentioned works considered only a single task, multi-task ML has recently been shown
to be advantageous, allowing to learn the metrics for several related tasks jointly
\cite{parameswaran2010large, yang2013geometry,yang2013multi}. Multi-task Large Margin Nearest
Neighbor (mt-LMNN) \cite{parameswaran2010large}, which is an extension of the (single task) LMNN
method~\cite{weinberger2009distance}, was one of the earliest multi-task ML methods. Given $T$
related tasks, mt-LMNN learns $T+1$ Mahlanobis-like metrics parametrized by matrices $M_0,
\{M_t\}_{t=1}^T$. 

$M_0$ encodes the general information common to all tasks while $M_t$'s encode the task specific
information. Since a full rank matrix is learned, the method scales poorly with feature dimensions.
Pre-processing with unsupervised compression techniques such as PCA is usually required, which
potentially leads to loss of information beforehand. Similarly, Wang \etal~\cite{wang2012multi}
proposed a multi-feature multi-task learning approach inspired by mt-LMNN. In general, mt-LMNN
suffers from overfitting. To overcome overfitting, Yang \etal~\cite{yang2013geometry} proposed a
regularizer based on Bregman matrix divergence~\cite{dhillon2007matrix}. In contrast with these
works, Yang \etal~\cite{yang2013multi} proposed a different but related approach aiming at learning
projection matrices $L_t \in \mathbb{R}^{d \times D}$ with $d\ll D$. They factorized these matrices
as $L_t = R_t^\top L_0$, where $L_0$ is common transformation matrix for all the tasks and $R_t$ are
task specific matrices. Their method is an extension of the Large Margin Component Analysis
(LMCA)~\cite{torresani2007LMCA}. It is important to note that LMCA requires $k$-nearest neighbors
for every classes in their objective function, and hence does not allow to handle tasks in which
only pairwise (dis-)similarity constraints are available. Furthermore, computing  the
$k$-nearest neighbors is computationally expensive.

In contrast to the works exploiting related tasks, Romera-Paredes \etal~\cite{romera2012exploiting}
proposed a multitask learning method which utilises a set of unrelated tasks, enforcing via
constraints that these tasks must not share any common structure.  Similarly, Du
\etal~\cite{du2015cross} used age verification as an auxiliary task to select discriminative
features for face verification. They use the auxiliary task to remove age sensitive features, with
feature interaction encouraged via an orthogonal regularization. Other works such
as~\cite{jayaraman2014decorrelating, liu2014feature, pu2014looks} discourage the sharing of features
between the unrelated set of tasks.

The application considered in this paper, \ie  face retrieval, requires encoding face images by
visual descriptors. This is another problem, widely addressed by the literature. Many different and
successful face features have been proposed such as  \cite{HussainBMVC2012, ojala2002, SharmaECCV12,
TanTIP10}. In the present work, we use signatures based on (i) Local Binary Patterns (LBP)
\cite{ojala2002} which are very fast to compute and have had a lot of success in face and texture
recognition,  and (ii) Convolutional Neural Networks (CNN) \cite{krizhevsky2012imagenet} which have
been shown to be very effective for face matching~\cite{TaigmanCVPR2014}. The computation of face
signatures is usually done after cropping and normalizing the regions of the images corresponding to
the faces. We do it by first locating face landmarks using the approach of Cao et al.
\cite{cao2014facelandmarks}.

\section{Approach}
\label{sec:approach}

As stated in the introduction, the proposed method aims at jointly learning  Mahalanobis-like
distances for $T$ different but related tasks, using positive and negative pairs from the different
tasks. The motivation is to exploit the relations between the tasks and potentially improve
performance.  
In such a case, the distance metric between vectors $\x_i,\x_j\in\R^D$ can be written as
\begin{align}
d_{M_t}^2(\x_i,\x_j) = (\x_i - \x_j)^\top M_t (\x_i - \x_j)
\end{align}
where $M_t\in\R^{D\times D}$ is a task specific parameter matrix (in the following, subscript $t$
denotes task $t$). To be a valid metric, $M$ must be positive semi-definite and hence can be
factorized as $M=L^\top L$. Following  \cite{MignonCVPR2012, weinberger2009distance} we decompose
$M$ as the square of a \emph{low rank} matrix $L \in \R^{d \times D}$, with $\mathrm{rank}(L) \leq d
\ll D$. This has the advantage that the distance metric can now be seen as a projection to a
Euclidean space of dimension $d \ll D$ \ie 
\begin{align} 
d_{\Lt}^2(\x_i,\x_j) = \| \Lt\x_i - \Lt\x_j \|^2,
\end{align}
thus resulting in a discriminative task-adaptive compression of the data. However, it has the
drawback that the optimization problem becomes non-convex in $L \ \forall d < D$, even if it was
convex in $M$ \cite{weinberger2009distance}. Nonetheless, it has been observed that even if  convergence
to global maximum is not guaranteed anymore, the optimization of this cost function is usually not
an issue and, in practice, very good results can be obtained \cite{GuillauminICCV2009, MignonCVPR2012}.

We consider an unconstrained setting with diverse but related tasks, coming from possibly different
heterogenous datasets. Training data consists of sets of annotated positive and negative pairs from
the different task related training sets, denoted as 
$\T_t = \{(\x_i, \x_j, y_{ij})\} \subset \R^D \times \R^D \times \{-1,+1\}$.
In the case of face matching, $\x_i$ and $\x_j$ are the face signatures while $y_{ij}=+1$ ($-1$)
indicates that the faces are similar (dissimilar) for the considered task \eg they are of the same
person (for identity retrieval) or they are of the same age (for age retrieval) or they both are
smiling (for expression retrieval). 

The main challenge here is to exploit the common information between the tasks \eg learning for age
matching might rely on some structure which is also beneficial for identity matching. Such
structures may or may not exist, as not only the tasks but also the datasets themselves are
different.

Towards this goal, we propose to couple the projections as follows: we define a generic global
projection $\Lz$ which is common for all the tasks, and, in addition, we introduce $T$ additional
task-specific projections $\{\Lt | t=1,\ldots,T\}$. The distance metric for task $t$ is then given
as
\begin{align}
    d^2_t(\x_i, \x_j )& = d^2_{L_0}( \x_i, \x_j ) + d_{\Lt}^2( \x_i, \x_j )  \nonumber \\
                                  & = \|L_0\x_i - L_0\x_j \|^2 + \|\Lt\x_i - \Lt\x_j \|^2.
    \label{eqnMultdist}
\end{align}
With this definition of $d_t$ we learn the projections $\{\Lz, L_1,\ldots,\Lt\}$ \emph{jointly} for
all the tasks.

\begin{algorithm}[t]
\begin{algorithmic}[1]
    \STATE \emph{Given}: $\{\T_t| t=1,\ldots,T\}, \eta_0, \eta$
\STATE \emph{Initialize}: $b_t=1$, $L_i \leftarrow\mathrm{wpca}(\T_i), L_0 \leftarrow L_1$
\FORALL{$i = 0,\ldots,$\texttt{niters}$-1$ }
\FORALL{$t = 0,\ldots,T-1$ }
    \IF{$ \mathrm{mod}(i, T) == t $}
        \STATE Randomly sample $(\x_i,\x_j, y_{ij})\in \T_t$ 
        \STATE Compute $d_t^2(\x_i,\x_j)$ using Eq.~\ref{eqnMultdist}
        \IF{$y_{ij}(b_t - d_t^2(\x_i,\x_j)) < 1$}
                \STATE $ L_0 \leftarrow L_0 - \eta_0 y_{ij}L_0(\x_i - \x_j)(\x_i - \x_j)^{\top}$
                \STATE $ L_t \leftarrow L_t - \eta y_{ij}L_t(\x_i - \x_j)(\x_i - \x_j)^{\top}$
                \STATE $b_t \leftarrow b_t + 0.1 \times \eta y_{ij} $ 
        \ENDIF
    \ENDIF
\ENDFOR
\ENDFOR
\caption{SGD for proposed CP-mtML}
\label{algo:ml}
\end{algorithmic}
\end{algorithm}

Learning the parameters of our CP-mtML model, \ie the projection matrices $\{\Lz, L_1,\ldots,\Lt\}$, is done by minimizing the total pairwise hinge loss given by:
\begin{align}
    \underset{\Lz,\{L_t,b_t\}_{t=1}^T}{\operatorname{argmin}} 
    \ \ \sum_{t=1}^T \sum_{\T_t} [1 - y_{ij}( b_t - d^2_t(\x_i, \x_j) )]_+ ,
    \label{our_app_obj}
\end{align}
with $[a]_+ = \max(0,a)$, $b\in\R$ being the bias, for all training pairs from all tasks. We optimize this function jointly \wrt all the projections, ensuring information sharing between the different tasks.

In practice, stochastic gradient descent (SGD) is used for doing this optimization. In each
iteration, we randomly pick a pair of images from a task, project them in (i) the common and (ii)
the corresponding task specific spaces and then compute the square of the Euclidean distance between
image descriptors in the respective sub-spaces. If the sum of distances violates the true
(dis-)similarity constraint, we update both matrices. To update the matrices, we use the closed-form
expression of the partial derivatives of the distance function $d_t$ \wrt $L_0, \Lt$, given by
\begin{align}
\frac{\pd d^2_t(\x_i, \x_j)}{\pd L_k} & =  L_k(\x_i - \x_j)(\x_i - \x_j)^{\top} \forall k = 0,\ldots,T
\label{eqn:grad}
\end{align}
%
%
Alg.~\ref{algo:ml} summarizes this learning procedure. 

The learning rates of the different projections are set as explained in the following. Regarding the
update of the common projection matrix, we can note that the update is done for every violating
training example of every task, while other projection matrices are updated much less frequently.
Based on this observation, the learning rate for task specific projection matrices is set to a
common value denoted as $\eta$ while the learning rate for the common projection matrix, denoted as
$\eta_0$, is set as a fractional multiple of $\eta$ \ie $\eta_0 = \gamma \eta$, where, $\gamma \in
[0,1]$ is a hyper-parameter of the model. The biases $b_t$ are task specific and are the 
thresholds on the distances separating positive and negative pairs. 
\subsection*{Advantage over mt-LMNN \cite{parameswaran2010large} }
The proposed distance function (Eq.~\ref{eqnMultdist}) can be rearranged and written as 
$d^2_t(\x_i, \x_j ) = ( \x_i - \x_j )^\top (L_0^\top L_0 + L_t^\top L_t) ( \x_i - \x_j )$
and thus bears resemblance to the distances learned with mt-LMNN  \cite{parameswaran2010large},
where $d^2_t(\x_i, \x_j ) = ( \x_i - \x_j )^\top (M_0 + M_t) ( \x_i - \x_j )$.  However, the
proposed model as well as the learning procedure are significantly different from
\cite{parameswaran2010large}. First, the objective function of mt-LMNN is based on triplets (while
our is based on pairs) \ie after projection a vector should be closer to another  vector of the same
class than to a vector of a different class. The learning procedure of mt-LMNN requires triplets
which is in general more difficult to collect and annotate than pairs. Second, despite the fact
that mt-LMNN leads to a semidefinite program which is convex, the proposed model has many practical
advantages. Since a low rank projection is learnt, there is no need for an explicit regularization
as limiting the rank acts as a regularizer. Another advantage is that the low dimensional
projections lead to a discriminative task-adaptive compression, which helps us do very efficient
retrieval. Third, the proposed SGD based learning algorithm is highly scalable and can work with
tens of thousands of examples in thousands of dimensional spaces, without any
compression/pre-processing of the features.  Finally, another big advantage of our approach is that
it can work in an online setting where the data streams with time.

\section{Experimental Results} 
\label{sec:experiments}
We now report the experiments we conducted to validate the proposed method for the task of face
retrieval based on traits which can be inferred from faces, including identity, age and expressions.
Such a task constitutes an important application domain of face based visual analysis methods. They
find application in security and surveillance systems as well as searching large human centered
image collections. In our experiments we focus on the two main tasks of identity and age based face
retrieval. For the former, we use age and expressions prediction tasks as auxiliary tasks while for
the later, we use identity prediction as the auxiliary task. We also evaluate identity based
retrieval at a very large scale, by adding a million of distractor faces collected independently
from the web. 

We now give details of the datasets we used for the evaluation, followed by the features and
implementation details and then discuss the results we obtain.
\vspace{0.5em} \\
\textbf{CASIA Web}~\cite{yi2014learning} 
 dataset consists of 494,414 images with weak annotations for 10,575 identities. We use this dataset to train
Convolutional Neural Network (CNN ) for faces. 
\vspace{0.5em} \\
\textbf{Labeled Faces in the Wild (LFW)}~\cite{huang2007labeled} 
is a standard benchmark for faces, with more than 13,000 images and around 5,000 identities. 
\vspace{0.5em} \\
\textbf{MORPH(II)}~\cite{ricanek2006morph} 
is a benchmark dataset for age estimation.  It has around 55,000 images annotated with both age and
identity. There are around 13,000 identities, with an average of 4 images per person, each at
different ages. We use a subset of around 13,000 images for our experiment.  We use this dataset for
age matching across identities and hence randomly subsample it and select one image per identity.
\vspace{0.5em} \\
\textbf{FACES}~\cite{ebner2010faces} 
is a dataset of facial expressions with 2052 images of 171 identities. Each identity has 6 different
expressions (neutral, happy, angry, in fear, disgusted, and sad) with 2 images of each. 
Here again, we sample one image from each of the expression of every person, and carefully avoid identity 
based pairings.
%
\vspace{0.5em} \\
\textbf{SECULAR}~\cite{bhattarai2014some} is a dataset  having one million face images extracted from Flickr. These are randomly crawled images and these images are not biased to any of the tasks or datasets. We use these images
as distractors during retrieval.

\subsection{Implementation details}
All our experiments are done with  grayscale images. The CNN model (details below) is trained with  normalized images of CASIA dataset. We use Viola and Jones~\cite{viola2004robust} face detector
for other datasets. For detecting facial key points and aligning the faces, we use the publicly
available implementation\footnote{\tiny \url{https://github.com/soundsilence/FaceAlignment}} of the facial 
keypoints detector of \cite{cao2014facelandmarks}. Faces are encoded using the following two features.
\vspace{0.5em} \\
\textbf{Local Binary Patterns (LBP).} We use the publicly available
$\texttt{vlfeat}$~\cite{Vedaldi2008} to compute descriptors. We resized the aligned face images to
$250\times250$ and centre cropped to $170\times100$.  We set cell size equal to $10$ for a descriptor of dimension $9860$.
\vspace{0.5em} \\
\textbf{Convolutional Neural Networks (CNN).} 
We use model trained on CASIA dataset with the architecture of Krizhevsky \etal
\cite{krizhevsky2012imagenet} to compute the feature of faces. Before computing the features, the
images are normalized similar to CASIA. We use the publicly available Caffe~\cite{jia2014caffe} deep 
learning framework  to train the model. The weights of the \texttt{fc7} layer are taken as the features 
($4096$ dimensions) and are $\ell_2$ normalized. As a reference, our features give a
verification rate of $88.4\pm1.4$ on the LFW dataset with unsupervised training setting ($+10\%$
compared to Fisher Vectors (FV) \cite{Simonyan13}) and $92.9\pm1.1$ with supervised metric learning
with heavy compression ($4096$ dimensions to $32$ dimensions) \cf $91.4\%$ for $16\times$ longer FVs.

\subsection{Compared methods.} 
We compared with the following three challenging methods for discriminative compression, using the
same features, same compressions and same experimental protocol for all methods for a fair
comparison.
\vspace{0.5em} \\
\textbf{WPCA} has been shown to be very competitive method for facial analysis -- even comparable to
many supervised methods \cite{HussainBMVC2012}. We compute the Whitened PCA from randomly sampled
subset of training examples from the main task. 
\vspace{0.5em} \\
\textbf{Single Task Metric Learning (stML)} learns a discriminative low dimensional projection for
each of the task independently. In Alg.~\ref{algo:ml}, we only have a global projection, with no
tasks, \ie $T=0$, reducing it to single task metric learning which we use as a baseline.
This is one of the state-of-art stML methods~\cite{Simonyan13} for face verification.
\vspace{0.5em} \\
\textbf{Metric Learning with Union of Tasks (utML). }
We also learn a metric with the union of all tasks to verify that we need different metrics for
different tasks instead of a global metric. We take all pairwise training data from all tasks and
learn a single metric as in stML above.
\vspace{0.5em} \\
\textbf{mtLMNN.} We did experiments with publicly available code of \cite{parameswaran2010large} but
obtained results only slightly better than WPCA and hence do not report them.
%

\subsection{Experimental Protocol}
We report results on two semantic face retrieval tasks, (i) identity based face retrieval and (ii)
age based face retrieval. We now give the details of the experimental protocol \ie details of metric
used, main experiments and how we create the training data for the tasks.
\vspace{0.5em} \\
\textbf{Performance measure.}
We report the \ock metric averaged over all the queries. $n$-call@$K \in [0,1]$ is an information
retrieval metric \cite{chen2006less} which is $1$ when at least $n$ of the top $K$ results retrieved
are relevant.  With $n=1$, this metric is relevant for evaluating real systems, \eg in security
and surveillance applications, where at least one of the top scoring $K$ retrievals should be the
person of interest, which can be further validated and used by an actual operator.
\vspace{0.5em} \\
\textbf{Identity based retrieval.} We use the LFW as the main dataset for identity based retrieval
experiments and MORPH (for age matching) and FACES (for expressions matching) as the auxiliary
datasets. We use $10,000$ (positive and negative) training pairs from LFW, disjoint from the query
images. For auxillary tasks, of expression and age matching, we randomly sample $40,000$ positive and
negative pairs, each. This setting is used to demonstrate performance improvements, when the data
available for auxiliary task is more than that for the main task. To compare our identity retrieval
performance with existing state-of-art rank boosting metric learning \cite{negrel2015boosted}, we
randomly sampled $25,000$ positive and negative pairs (\cf $\sim 32,000$ by
\cite{negrel2015boosted}) and take the same sets of constraints as before from auxiliary tasks. 

Following Bhattarai \etal~\cite{bhattarai2014some}, we choose one random image from the identities
which have more than five images, as query images and the rest as training images. This gives us 423
query images in total. We use these images to do Euclidean distance, in the projection space, based
nearest neighbor retrieval from the rest of the images, one by one. The non-query images are used to
make identity based positive and negative pairs for the main task. We use two auxiliary tasks, (i)
age matching using MORPH and (ii) expressions matching using FACES.
\vspace{0.5em} \\
\textbf{Age based retrieval.} We use the MORPH dataset as the main dataset and the LFW dataset as
the auxiliary dataset. We randomly split the dataset into two disjoint parts as train+validation and
test sets. In the test set, one image from each age class is taken as the probe query while the rest
make the gallery set for retrieval. We take $10,000$ age pairs and $30,000$ of identity pairs. 
\vspace{0.5em} \\
\textbf{Large scale retrieval with 1M distractors.} We use the SECULAR dataset for distractors. We
make the assumption that, as these faces are crawled from Flickr accounts of randomly selected
common users, they do not have any identity present in LFW, which is a dataset of famous people. With
this assumption, we can use these as distractors for the large scale identity based retrieval task
and report performances with the annotations on the main dataset, since all of the distractors will
be negatives. However, we can not make the same assumption about age and hence we do not use
distractors for age retrieval experiments.
\vspace{0.5em} \\
\textbf{Parameter settings.}
We choose the values for the parameters ($\eta, \eta_0$, \texttt{niters}) by splitting the train set
into two parts and training on one and validating on the other \ie these sets were disjoint from all
of the test sets used in the experiments. 

\subsection{Quantitative Results}

We now present the quantitative results of our experiments. We first evaluate the contributions of
the different projections learnt, \ie the common projection $\Lz$ and the task specific projection $\Lt$,
in terms of performance on the main task. We then show the performance of the proposed CP-mtML \wrt
the compared methods on the two experiments on (i) identity based and (ii) age based face retrieval. 
We mention the auxiliary task in brackets \eg CP-mtML (expr) means that the auxiliary task was
expression matching, with the main task being clear from context.
\begin {table}[t]
\newcolumntype{L}[1]{>{\raggedright\let\newline\\\arraybackslash\hspace{0pt}}m{#1}}
\newcolumntype{C}{>{\centering\arraybackslash}p{3.5em}}
\newcolumntype{D}{>{\centering\arraybackslash}p{2.5em}}
\centering
\begin{tabular}{ | c | C | D | D | D |}
\hline
Projection &  $K=2$       &  $5$   & $10$  & $20$  \\
\hline
\hline
$L_0$                          &  30.3  & 38.1  & 43.3  & 51.8  \\
\hline
$L_1$                          &  35.0   & 46.6  & 55.8  & 64.8 \\
\hline
$L_2$                          &  4.5   & 7.6   & 10.4  & 13.0 \\
\hline
$L_0+L_1$                      & \bf{43.5}   & \bf{55.6}  & \bf{63.6}  & \bf{69.5} \\
\hline
\end{tabular}
\caption{ Performance (\ock) of different projections matrices learned with proposed CP-mtML (LBP
features, $d=64$) for identity retrieval with auxiliary task of expression matching.
}
\label{tabProjMix}
\vspace{-0.8em}
\end {table}
\vspace{0.5em} \\
\textbf{Contributions of projections.} Tab.~\ref{tabProjMix} gives the performance of the different 
projections for the task of identity based retrieval task with expression matching as the auxiliary
task. We observe an expected trend; the combination of the common projection $L_0$ with the task specific
one $L_1$ performs the best at $69.5$ at $K=20$. The projection for the auxiliary task $L_2$ expectedly does
comparatively badly at $13.0$, as it specializes on the auxiliary task
and not on the main task. The projection $L_1$ specializing on the main task is better than the
common projection $L_0$ ($64.8$ \vs $51.8$) while their combination is the best ($69.5$). The trend
was similar for the auxiliary task. This demonstrates that the projection learning follows the
expected trend, the global projection captures commonalities and in combination with the task
specific projections performs better for the respective tasks.
\begin{table*}
\centering
\newcolumntype{C}{>{\centering\arraybackslash}p{2.9em}}
\newcolumntype{K}{>{\centering\arraybackslash}p{5.2em}}
\begin{tabular}{|K|K|CCCC|CCCC|}
    \cline{3-10}
                \multicolumn{2}{c|}{\ } & \multicolumn{4}{c|}{No distractors}   &
                \multicolumn{4}{c|}{1M distractors} \\
    \hline
    Method  &   Aux    & $K=2$     & $5$    & $10$   & $20$                 & $K=2$    & $5$ & $10$        & $20$ \\
    \hline
    \hline
    WPCA    &   n/a      & 30.0   & 37.4  & 43.3  & 51.3                    & 24.6    & 28.8       & 33.8       & 39.0 \\
    stML    &   n/a      & 38.1   & 51.1  & 60.5  & 69.3                    & 26.0     & 37.4       &43.3        & 48.7 \\
    \hline
    utML    &  expr & 31.0    &38.1   & 48.5  & 57.9                        & 20.3    &25.8        &31.9        & 38.5 \\
    CP-mtML &  expr & \bf{43.5}   &\bf{55.6}   &\bf{63.6}   & \bf{69.5}     & \bf{33.1}    &\bf{43.3}        &\bf{51.1}        &\bf{55.3}    \\
    \hline
    utML    &  age     & 21.7   & 31.4  & 41.1  & 53.0                      & 12.8   & 18.9       & 24.6      & 31.7           \\
    CP-mtML &  age     & \bf{46.1}   &\bf{56.0}   &\bf{63.4}  & 68.3   & \bf{35.7}   & \bf{43.5}      & \bf{47.8}      & \bf{52.2} \\
    \hline
\end{tabular}
\caption{Identity based face retrieval performance (1-call@$K$ for different $K$) with and without distractors with LBP features. 
Auxiliary task is either Age or Expression matching. Projection dimension, $d=64$} 
\label{tabIdLBP}
\vspace{-0.3em}
\end{table*}
\begin{table*}
\centering
\newcolumntype{C}{>{\centering\arraybackslash}p{2.9em}}
\newcolumntype{K}{>{\centering\arraybackslash}p{5.2em}}
\begin{tabular}{|K|K|CCCC|CCCC|}
    \cline{3-10}
                \multicolumn{2}{c|}{\ } & \multicolumn{4}{c|}{No distractors}   &
                \multicolumn{4}{c|}{1M distractors} \\
    \hline
    Method  &   Aux      & $K=2$     & 5     & 10   & 20                & $K=2$      & 5         & 10        & 20 \\
    \hline
    \hline
    WPCA    &   n/a      & 72.1   & 80.4  & 83.7  & 89.1                           & 65.2    & 72.1       & 75.9       & 78.7 \\
    stML    &   n/a      & \bf{76.8}   & 85.1  & 89.6  & 92.0                         & 70.7    & 78.0       & 82.0       & 84.2 \\
    \hline
    utML    &  expr & 73.5   &82.3   &87.2   & 90.3                                  & 67.1    & 76.8       & 79.0       & 82.0  \\
    CP-mtML    &  expr & \bf{76.8}   &\bf{86.5}  &\bf{90.3} &\bf{93.4}                  & \bf{71.2} &\bf{79.7} &\bf{83.2}   &\bf{85.3}     \\
    \hline
    utML     &  age     & 73.0   & 82.0  & 88.2  & 91.0                                  & 68.1    &76.1        & 81.1       & 82.7           \\
    CP-mtML    &  age     & \bf{76.8} & \bf{85.8} & \bf{90.3} & \bf{93.6}                  & \bf{71.2}  & \bf{79.0} & \bf{83.0} & \bf{85.1} \\
    \hline
\end{tabular}
\caption{Identity based face retrieval performance (1-call@$K$ for different $K$) with and without distractors with CNN features.
Auxiliary task is either Age or Expression matching. Projection dimension, $d=64$} 
\vspace{-.5em} 
\label{tabIdCNN}
\end{table*}
\begin{table*}
\newcolumntype{L}[1]{>{\raggedright\let\newline\\\arraybackslash\hspace{0pt}}m{#1}}
\newcolumntype{C}[1]{>{\centering\arraybackslash}p{#1}}
\centering
\begin{minipage}{0.56\textwidth}
\resizebox{\linewidth}{!} {
\begin{tabular}{|c|c|ccc|ccc|}
    \cline{3-8}
                \multicolumn{2}{c|}{\ }      &
                \multicolumn{3}{c|}{No distractor}   &
                \multicolumn{3}{c|}{1M distractors} \\
    \hline
    Method  &   Aux     & $d=32$       & $64$         & $128$                             & $d=32$      & $64$         & $128 $       \\
    \hline
    \hline
    WPCA    &   -       & 34.3       & 43.3         & 52.5                              & 23.4      & 33.8         & 40.4       \\
    stML    &   -       & 50.1       & 60.5         & 63.6                              & 33.3      & 43.3         & 51.3         \\
    \hline
    utML    &  expr  & 44.2       &48.5          & 57.4                              & 25.3    &31.9        &31.9       \\
    CP-mtML    &  expr  & \bf{55.6}  &\bf{63.6}     &\bf{70.2}                          & \bf{37.6}    &\bf{51.1}        &\bf{54.6}       \\
    \hline
    utML    &  age      & 37.6       & 41.1         & 51.5                              & 17.5      & 24.6         & 34.0               \\
    CP-mtML    &  age      & \bf{52.5}  &\bf{63.4}     &\bf{69.0}                          & \bf{34.3} & \bf{47.8}    & \bf{53.9}      \\
    \hline
\end{tabular}
}
\end{minipage}
\begin{minipage}{0.408\textwidth}
\newcolumntype{L}[1]{>{\raggedright\let\newline\\\arraybackslash\hspace{0pt}}m{#1}}
\newcolumntype{C}[1]{>{\centering\arraybackslash}p{#1}}
\centering
\resizebox{\linewidth}{!} {
\begin{tabular}{|ccc|ccc|}
    \hline
                \multicolumn{3}{|c|}{No distractor}   &
                \multicolumn{3}{c|}{1M distractors} \\
    \hline
    $d=32$       & $64$         & $128$                             & $d=32$      & $64$         & $128 $       \\
    \hline
    \hline
    83.9        & 83.7         & 85.6                             & 74.5         & 75.9         & 75.2       \\
    88.4        & 89.6        & 88.7                              & 80.6         & 82.0         & \bf{81.6}         \\
    \hline
     85.1        &87.2          & 86.3                              & 73.0         &79.0          &78.3       \\
    \bf{88.7}    &\bf{90.3}     &\bf{89.4}                          & \bf{81.3}    &\bf{83.2}     & 81.1       \\
    \hline
    85.3       & 88.2          & 86.5                             & 76.6         & 81.1         & 79.2               \\
    88.2       &\bf{90.3}      &\bf{89.6}                          & \bf{80.9}    & \bf{83.0}    & \bf{81.6}      \\
    \hline
\end{tabular}
}
\end{minipage}
\caption{Identity based face retrieval, 1-call@10 at different projection dimension, $d$, (left)
using LBP and (right) CNN features.} 
\vspace{-.5em} 
\label{tabIdProjDim}
\end{table*}
\vspace{0.5em}\\
\textbf{Identity based retrieval.}
We evaluate identity based face retrieval with two different features \ie LBP and CNN, both with and
without one million distractors. Tab.~\ref{tabIdLBP} and \ref{tabIdCNN} give the performances of the
different methods for different values of $K$ (the number of top scoring images considered). First
of all we notice the general trend that the performances are increasing with $K$, which is expected.
We see that, both in the presence and absence of distractors, the proposed method performs
consistently the best compared to all other methods. In the case of LBP features, the performance
gains are slightly more when the auxiliary task is age prediction \eg $46.1$ for CP-mtML (age) \vs
$43.5$ for CP-mtML (expr) at $K=2$, both these values are much better than WPCA and stML ($30.0$ and
$38.1$ ) respectively. Interestingly, when we take all the tasks together and learn only a single
projection, \ie utML, it degrades for both age and expression as auxiliary tasks, but more so for age
($21.7$ \vs $31$). This happens because the utML projection brings similar age people closer and
hence confuses identity more, as age is more likely to be shared compared to expressions which are
characteristic of different people. The proposed CP-mtML is not only able to recover this loss but
also leverages the extra information from the auxiliary task to improve performance of the main
task.

When distractors are added the performances generally go down \eg $68.3$ to $52.2$ for LBP and
$93.6$ to $85.1$ for CNN with CP-mtML (age). However, even in the presence of distractors the
performance of the proposed CP-mtML is better than all other methods, particularly stML \eg $43.3$
for CP-mtML (expr) \vs $37.4$ for stML at $K=5$ with LBP and $79.7$ for CP-mtML (expr) \vs $78.0$
for stML with CNN.

The performances of the two different features are quite different. The lightweight unsupervised LBP
features perform lower than the more discriminative CNN features, which are trained on large amounts
of extra data \eg $86.5$ \vs $55.6$ at $K=5$ for CP-mtML (expr). The performance gains for the
proposed method are larger for LBP compared to CNN features \eg $+4.5$ \vs $+1.4$ at $K=5$ for
CP-mtML (expr) \cf stML. While such improvements are modest for CNN features, they are consistent
for all the cases. Parallely, the improvements for LBP features are substantial, especially in the
presence of distractors \eg $+7.8$ for CP-mtML (expr) \vs stML at $K=10$. While it may seem that
using stronger feature should then be preferred over using a stronger model, we note that this may
not be always preferable. In a surveillance scenario, for instance, where a camera just records
hours of videos and we need to find a specific face after some incident, using time efficient
features as a first step for filtering and then using the stronger feature on a sufficiently small
set of filtered examples is advantageous. This is highlighted by the time complexities of the
features; in practice LBP are much faster than CNN to compute. While CNN features roughly take
$450$ milliseconds, the LBP features take only a few milliseconds on a $2.5$ GHz processor.

Further, Tab.~\ref{tabIdProjDim} presents the $1$-call@$10$ while varying the projection dimension,
which is directly proportional to the amount of compression. We observe that all methods gain
performance when increasing the projection dimension, however, with diminishing returns. In the
presence of one million distractors, CP-mtML (expr) improves by $+13.5$ when going from $d=32$ to
$d=64$ and $+3.5$ when going from $d=64$ to $d=128$ for LBP. The results for larger $d$ were
saturated for LBPs with a slight increase. The performance changes with varying $d$ in the presence
of distractors for CNN features are more modest. CNN with distractors gets $+1.9$ for $d=32$ to
$d=64$ and $-2.1$ for $d=64$ to $d=128$ \ie the algorithm starts over-fitting at higher dimensions
for the stronger CNN features. As an idea of space complexity, at compression to $d=32$ dimensional
single precision vector per face, storing ten million faces  would require one gigabytes of space,
after projection. Interestingly, the proposed method is better than stML in all but one case (CNN
features with $d=128$) which is a saturated case anyway.

Tab.~\ref{tabMLBOOST} gives the comparisons (with LBP features and $d=32$) with MLBoost
\cite{negrel2015boosted}. At $K=10$ CP-mtML obtains $61.5,58.9$ with age and expressions as
auxiliary taks, respectively, while the MLBoost method stays at $54.1$. Hence the proposed method is
better than the results reported in the literature. As said before, we also used the publicly available code of mtLMNN
\cite{parameswaran2010large}. We obtained results only slightly better than WPCA and hence do not
report them.

With the above results we conclude the following. The proposed method effectively leverages the
additional complementary information in the related tasks of age and expression matchings, for the
task of identity based face retrieval. It consistently improves over the unsupervised WPCA,
supervised stML which does not use additional tasks and also utML which combines all the data.
It is also better than these methods at a range of projection dimensions (\ie compressions),
deteriorating only at the saturated case of high dimensions with strong CNN features.

\begin{figure*}[t]
\centering
\includegraphics[width=0.32\textwidth, trim=0 40 220 0, clip]{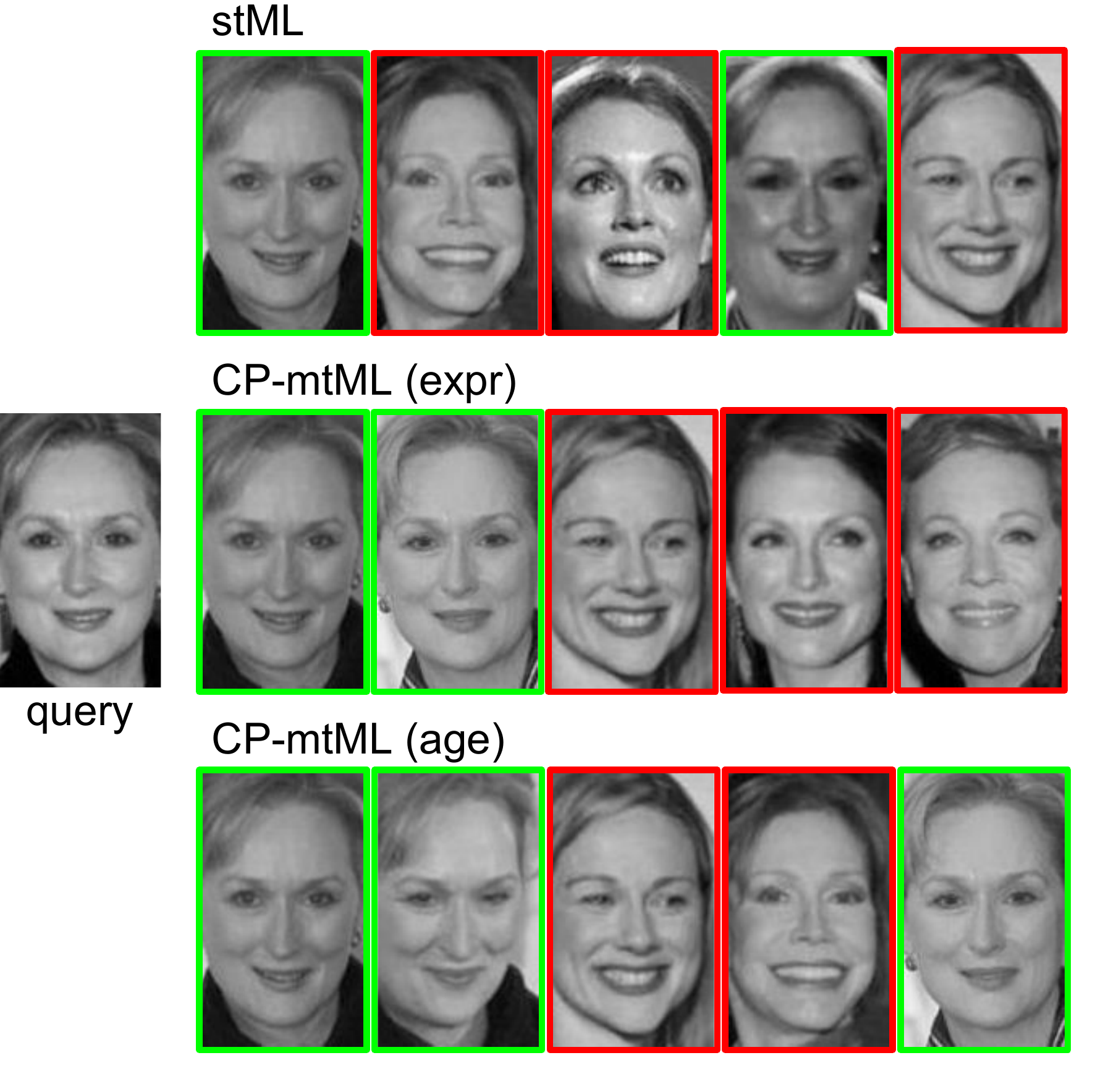} \hfill
\includegraphics[width=0.32\textwidth, trim=0 40 220 0, clip]{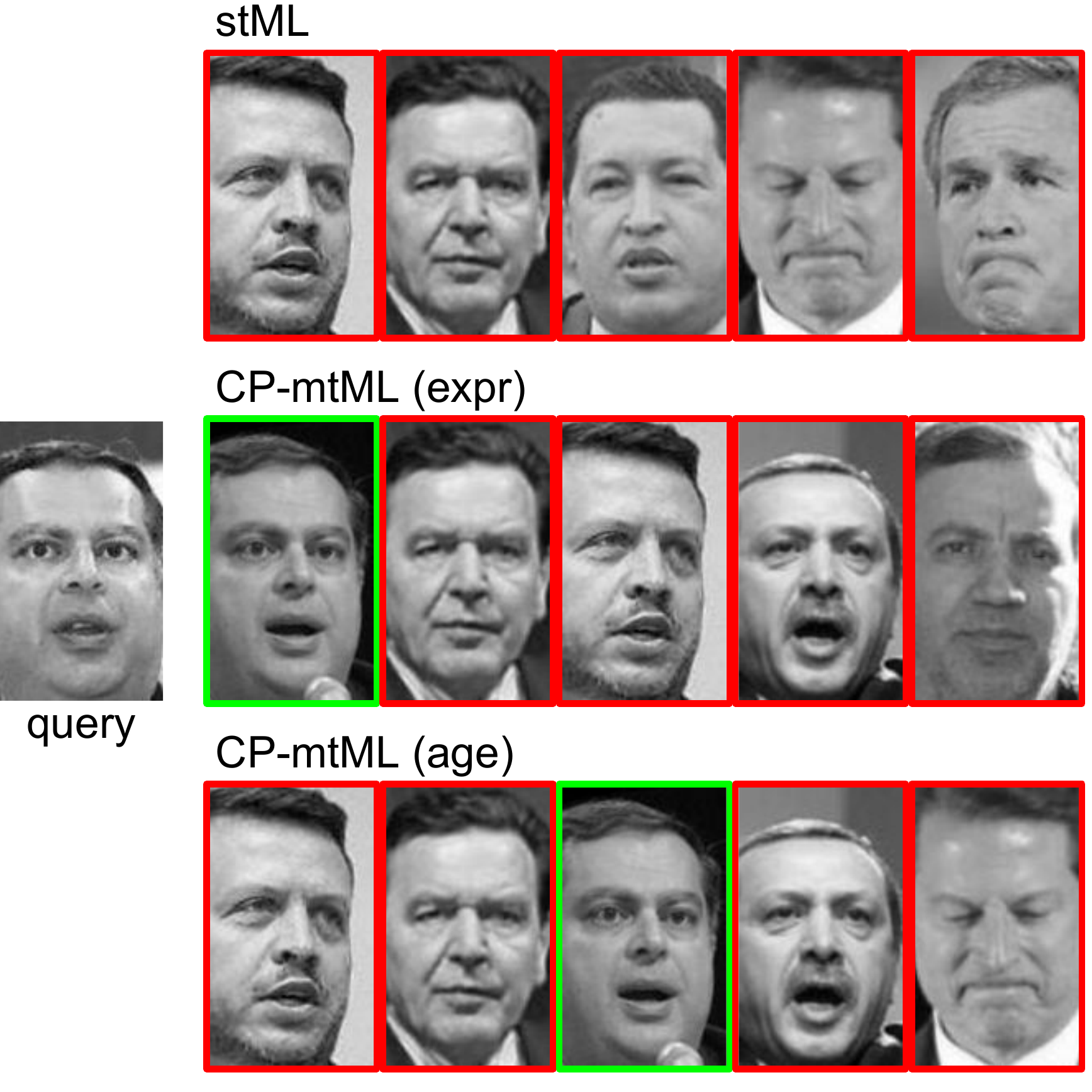} \hfill
\includegraphics[width=0.32\textwidth, trim=0 40 220 0, clip]{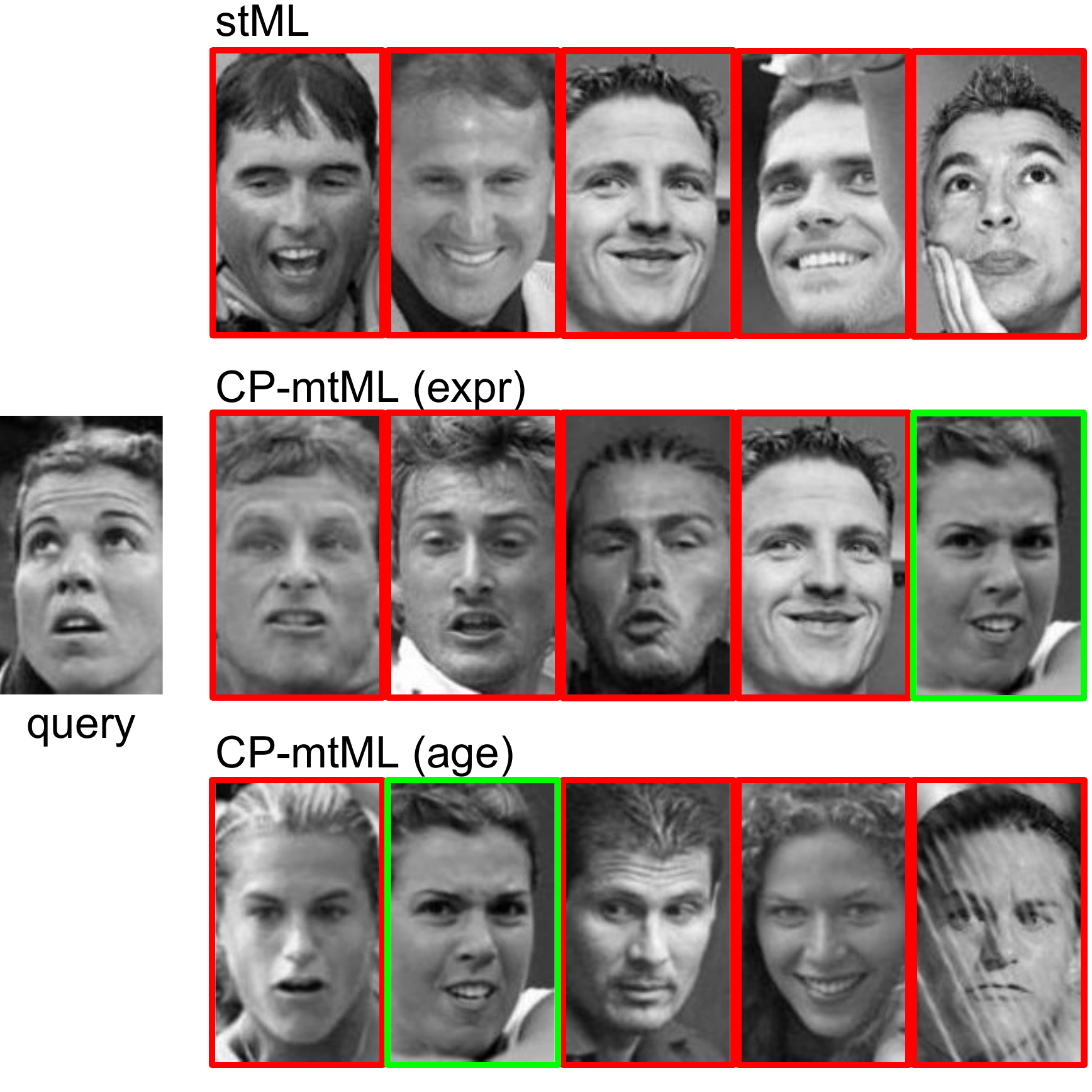}
\hfill
\caption{ 
    The $5$ top scoring images (LBP \& no distractors) for three queries for the different methods
    (auxiliary task in brackets). True (resp. False) Positive are marked with a green (resp. red) border (best viewed in color).
}
\label{figQualId}
\end{figure*}

\begin{table}
\centering
\newcolumntype{C}{>{\centering\arraybackslash}p{2.1em}}
\newcolumntype{K}{>{\centering\arraybackslash}p{5.2em}}
\begin{tabular}{|r|c|CC|CC|}
    \cline{3-6}
                \multicolumn{2}{c|}{\ } & \multicolumn{2}{c|}{\small No distractors}   &
                \multicolumn{2}{c|}{\small 1M distractors} \\
    \hline
    Method   &  Aux  & \small{$K$=10}  & \small{20} & \small{10} & \small{20} \\
    \hline
    \hline
    MLBoost         &   n/a       & 54.1        & 63.4             & 34.3       & 39.5 \\
    \hline
    CP-mtML         &  expr       & \bf{58.9 }       &\bf{69.5}              &\bf{38.1}       & \bf{45.6}     \\
    \hline
    CP-mtML         &  age         & \bf{61.5}         & \bf{70.7}            & \bf{39.7}        & \bf{47.8} \\
    \hline
\end{tabular}
\caption{Performance comparison with existing MLBoost~\cite{negrel2015boosted} (for LBP features and $d=32$).}
\vspace{-1em} 
\label{tabMLBOOST}
\end{table}

\begin{figure}[t]
\centering
\includegraphics[width=.95\columnwidth, trim=10 15 -10 5, clip]{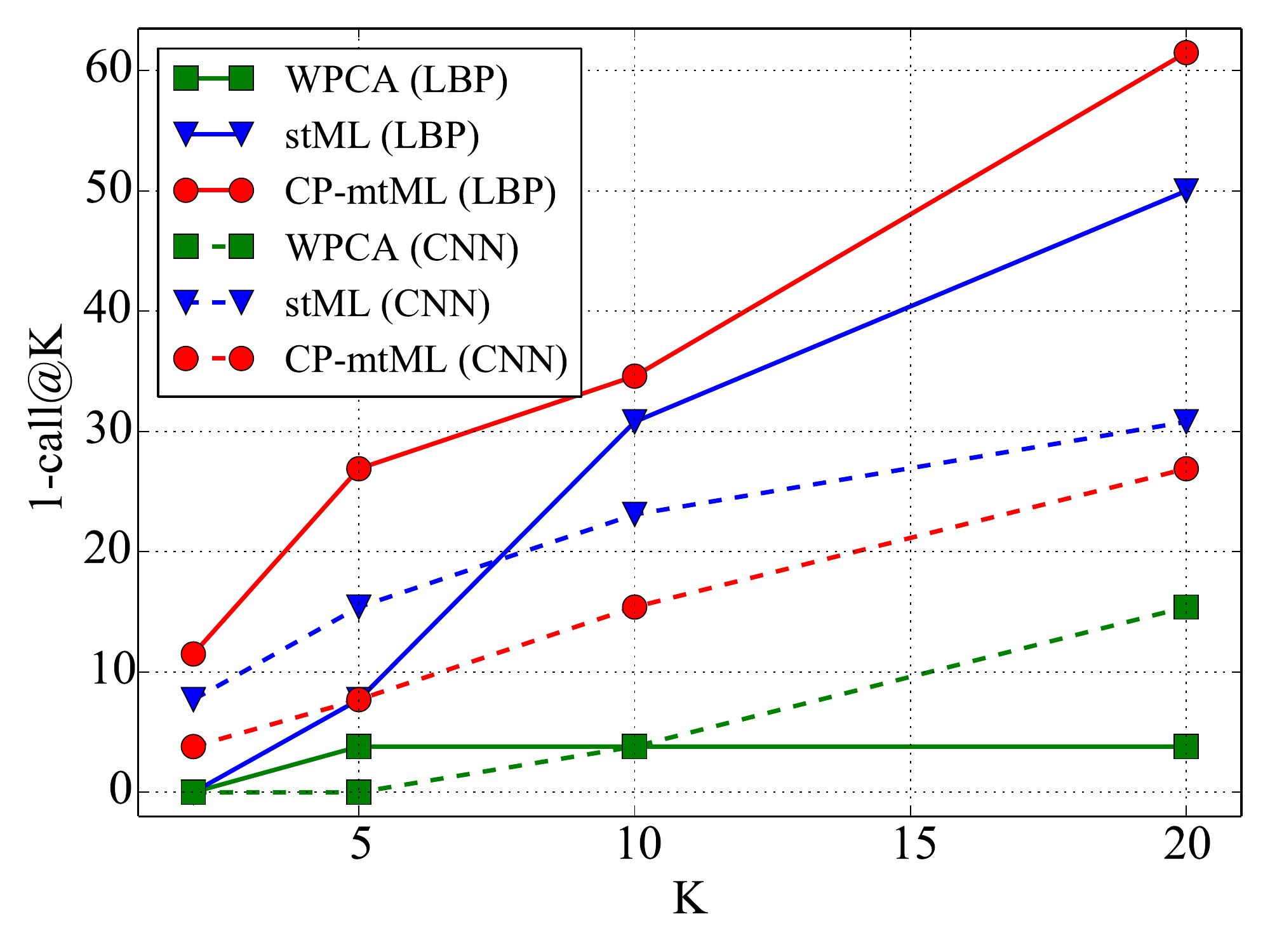}
\hfill
\caption{ 
Age retrieval performance (1-call@$K$) for different $K$ with auxiliary task of identity matching.
The dimension of projection is $d=32$}
\label{figAge}
\vspace{-1em} 
\end{figure}

\vspace{0.5em} \noindent
\textbf{Age based retrieval.} Fig.~\ref{figAge} presents some results for face retrieval based on
age for the different methods, with the auxiliary task being that of identity matching. 
In this task CP-mtML outperforms all other methods by a significant margin with LBP features.  These
results are different and interesting from the identity based retrieval experiments above, as they
show the limitation of CNN features, learnt on identities, to generalize to other tasks --- the
performances with LBP features are higher than those with CNN features. 

While the trend is similar for LBP features \ie CP-mtML is better than stML, it is reversed for CNN
features.  With CNN features, stML learns to distinguish between ages when trained with such data,
however, CP-mtML ends up being biased, due to its construction, towards identity matching and
degrades age retrieval performance when auxiliary task is identity matching. However, the
performance of CPmtML with LBP features is much higher than of any of the methods with CNN features.
\subsection{Qualitative results}
We now present some qualitative comparisons between the proposed CP-mtML, with age and expression
matching as auxiliary tasks, with the competitive stML method. Fig.~\ref{figQualId} shows the top
five retrieved faces for three different queries for stML and the proposed CP-mtML with age and
expression matching as auxiliary tasks. The results qualitatively demonstrate the better
performance obtained by the proposed method. In the first query (left) all the methods were able to
find correct matches in the top five. While stML found two correct matches at ranks $1$ and $4$,
CP-mtML (age) also found two but with improved ranks \ie $1$ and $2$ and CP-mtML (expression) found
three correct matches with ranks $1,2$ and $5$. While the first query was a relatively simple query,
\ie frontal face, the other two queries are more challenging due to non-frontal pose and
deformations due to expression. We see that stML completely fails in these cases (for $K=5$) while
the proposed CP-mtML is able to retrieve one correct image with ranks $1,3$ (middle) and $5,2$
(right) when used with age and expression matching as auxiliary tasks, respectively. 
It is interesting to note that with challenging pose and expression the appearances of the
faces returned by the methods are quite different (right query) which demonstrates that CP-mtML
projection differs
from that learned by stML.

\section{Conclusions}
\label{sec:conclusions}
We presented a novel Coupled Projection multi-task Metric Learning (CP-mtML) method for leveraging
information from related tasks in a metric learning framework. The method factorizes the information
into different projections, one global projection shared by all tasks and $T$ task specific
projections, one for each task. We proposed a max-margin hinge loss minimization objective based on
pairwise constraints between training data. To optimize the objective we use an efficient stochastic
gradient based algorithm. We jointly learn all the projections in a holistic framework which leads
to sharing of information between the tasks. We validated the proposed method on challenging tasks
of identity and age based image retrieval with different auxiliary tasks, expression and age
matching for the former and identity matching in the later. We showed that the method improves
performance when compared to competitive existing approaches. We analysed the qualitative results,
which also supported the improvements obtained by the method.

{\small
\bibliographystyle{ieee}
\bibliography{biblio}
}
\section{Additional Results}
In this section we present additional both quantiative
and qualitative results.
\subsection{Quantitative Results}
In this section, we compare performance of existing state-of-art multitask metric learning method, mtLMCA of Yang 
\etal ~\cite{yang2013multi} with the performance of the proposed method and other baselines. In addition to it, 
we present the in-depth analysis of the proposed algorithm such as it's time complexity and scalability. 
We then present the optimization curves of loss functions of our method and mtLMCA.

\vspace{1em} \noindent
\textbf{Comparisons with mtLMCA.} We implemented the existing mtLMCA and compare the performance with the proposed method.
For mtLMCA, we initialized the the common projection,~$L_0$ and task specific,~$R_t$ matrices with identity matrices as explained 
in the paper. Whereas, for rest of the cases, as stated in the 
Alg.~\ref{algo:ml} with the WPCA.
\begin{table*}
\centering
\newcolumntype{C}{>{\centering\arraybackslash}p{2.9em}}
\newcolumntype{K}{>{\centering\arraybackslash}p{5.2em}}
\begin{tabular}{|K|K|CCCC|CCCC|}
    \cline{3-10}
                \multicolumn{2}{c|}{\ } & \multicolumn{4}{c|}{No distractors}   &
                \multicolumn{4}{c|}{1M distractors} \\
    \hline
    Method  &   Aux    & $K=2$     & $5$    & $10$   & $20$                 & $K=2$    & $5$ & $10$        & $20$ \\
    \hline
    \hline
    WPCA    &   n/a      & 30.0   & 37.4  & 43.3  & 51.3                    & 24.6    & 28.8       & 33.8       & 39.0 \\
    stML    &   n/a      & 38.1   & 51.1  & 60.5  & 69.3                    & 26.0     & 37.4       &43.3        & 48.7 \\
    \hline
    utML    &  expr & 31.0    &38.1   & 48.5  & 57.9                        & 20.3    &25.8        &31.9        & 38.5 \\
    mtLMCA  &  expr & 29.3  &40.7   & 48.0  & 61.0                        & 19.9    &28.4        &34.8        & 40.0  \\
    CP-mtML &  expr & \bf{43.5}   &\bf{55.6}   &\bf{63.6}   & \bf{69.5}     & \bf{33.1}    &\bf{43.3}        &\bf{51.1}        &\bf{55.3}    \\
    \hline
    utML    &  age     & 21.7   & 31.4  & 41.1  & 53.0                      & 12.8   & 18.9       & 24.6      & 31.7           \\
    mtLMCA  &  age   & 27.4   & 39.7  & 50.4  & 61.0                      & 18.7   & 24.6       & 29.8      & 35.5         \\
    CP-mtML &  age     & \bf{46.1}   &\bf{56.0}   &\bf{63.4}  & 68.3   & \bf{35.7}   & \bf{43.5}      & \bf{47.8}      & \bf{52.2} \\
    \hline
\end{tabular}
\caption{Identity based face retrieval performance (1-call@$K$ for different $K$)  with and without distractors with LBP features. 
Auxiliary task is either Age or Expression matching. Projection dimension, $d=64$} 
\label{tabIdLBP2}
\vspace{-0.8em}
\end{table*}

Tab.~\ref{tabIdLBP2} shows the performance comparison. 
In comparison with mtLMCA, we observe that the proposed 
CP-mtML outperforms mtLMCA by a significant margin. We explain it as follows.
Without loss of generality consider task 1~(\eg identity matching), the projection by proposed method is given by a 
common $L_0$ and a task specific $L_1$ while that by mtLMCA is given by common $L_0$ and
task specific $R_1$ . While $L_0$ , $L_1$ are both $d \times D $ matrices $R_1$ is $d \times d$. Hence 
in CP-mtML there are $dD$ common (across tasks) parameters and $dD$ task specific parameters, while
mtLMCA has same $dD$ common parameters but only $d^2$ task specific parameters. We suspect that with equal number
of task specific and common parameters CP-mtML is able to exploit the shared as well as task specific information
well while for mtLMCA the small number of task specific parameters are not able to do so effectively e.g. for the specific 
case of 9860D LBP features projected to 64D, while 50\% of the parameters are task specific for CP-mtML, only
$64^2 /(9860 \times 64) = 0.7$\% are task specific in mtLMCA. In addition to it, we could see this method as utML with 
a very small fraction of task specific parameters. As mentioned before, utML learns a common projection matrix taking 
training examples from both the domains. From the performance also, it supports our argument. 
We can see that the performance of mtLMCA is slightly better than utML. This is due to the small separate task specific 
parameters in mtLMCA. Our proposed method, CP-mtML is capable of learning large task specific parameters 
maintaining the same projection dimension as that of other methods, which ultimately gives the improved performance. 
\begin{figure}[t]
\centering
\includegraphics[width=.95\columnwidth, trim=10 15 -10 5, clip]{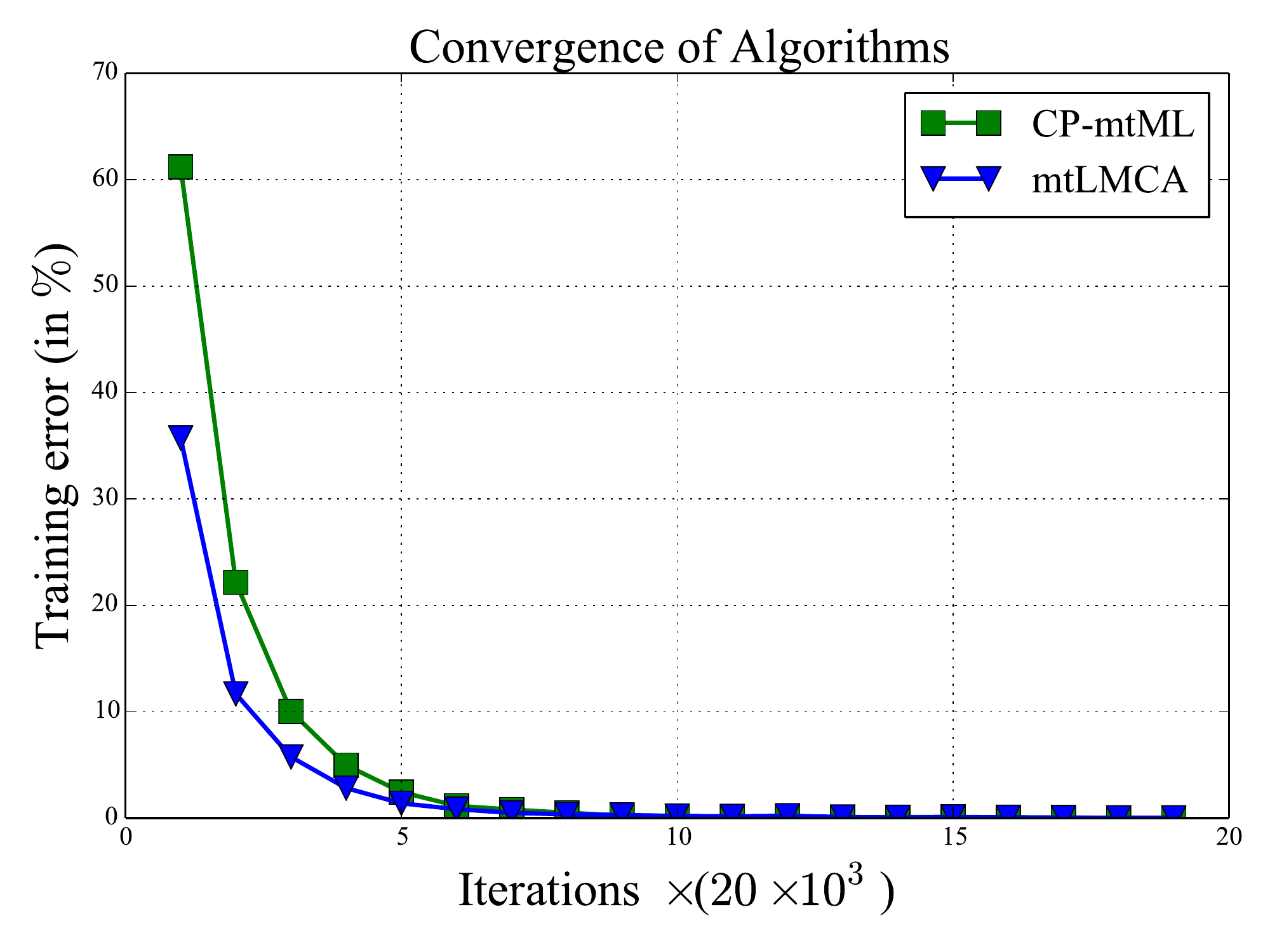}
\hfill
\caption{ 
Optimization of loss functions}
\label{figOpt}
\vspace{-1em} 
\end{figure}

\vspace{1em} \noindent
\textbf{Time Complexity and Scalability.}
CP-mtML is about 2.5$\times$ slower to train than stML -- specifically
it takes 40 minutes to train CP-mtML with 50, 000 training pairs while compressing 
9860D LBP features to 64D on a single core of 2.5 Ghz system running Linux. The training 
time is linear in the number of training examples. As
the 64D features are real vectors it takes 256 bytes (with
4 bytes per real) to index one face or about a manageable
1.8 TB to index the current human population of about 7
billion people; hence we claim scalability.

\vspace{1em} \noindent
\textbf{Convergences of Algorithms.}
Fig.~\ref{figOpt} shows the convergences of CP-mtML and mtLMCA. From the figure, we see that 
both the algorithms are converged well.

\subsection{Qualitative Results}

We present some more qualitative results to compare the proposed Coupled Projection multi-task
Metric Learning (CP-mtML) with the most competitive baseline \ie Single Task Metric Learning (stML).
The main task here is that of identity based face retrieval while the auxiliary tasks are expression
(expr) and age (age) based matching.

\begin{figure*}[t]
\centering
\includegraphics[width=0.32\textwidth, trim=0 40 200 0, clip]{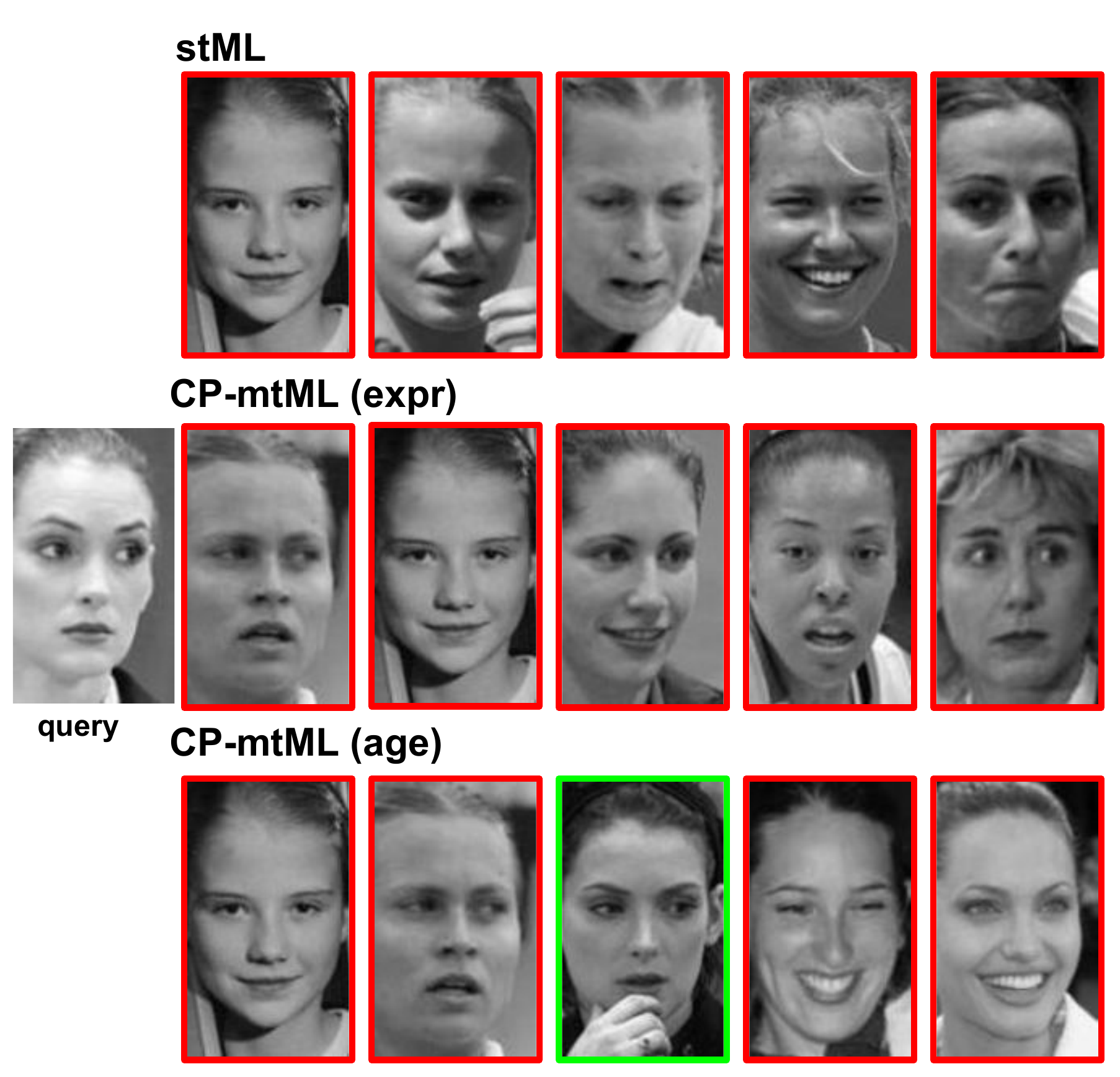} \hfill
\includegraphics[width=0.32\textwidth, trim=0 40 200 0, clip]{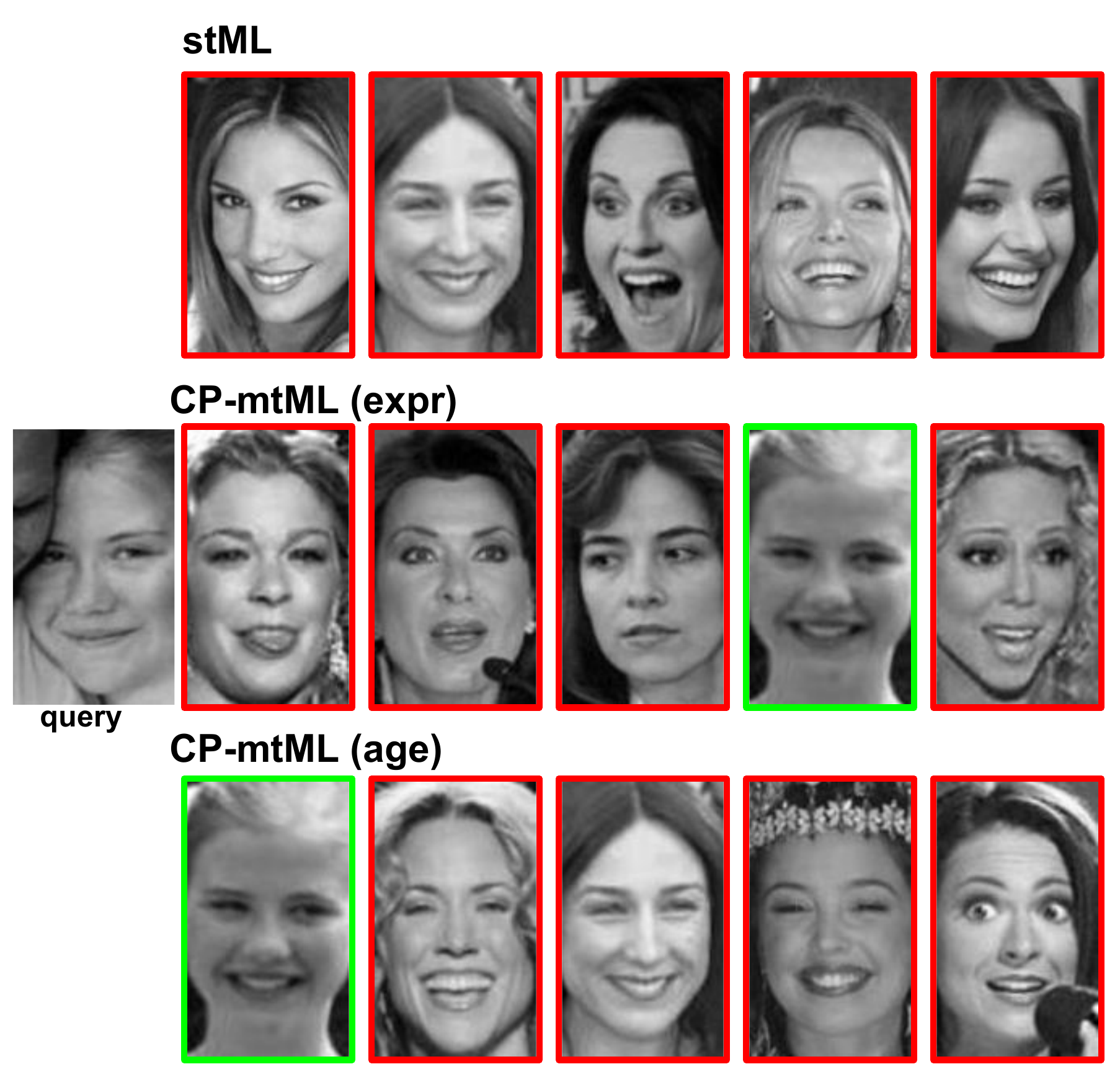} \hfill
\includegraphics[width=0.32\textwidth, trim=0 40 200 0, clip]{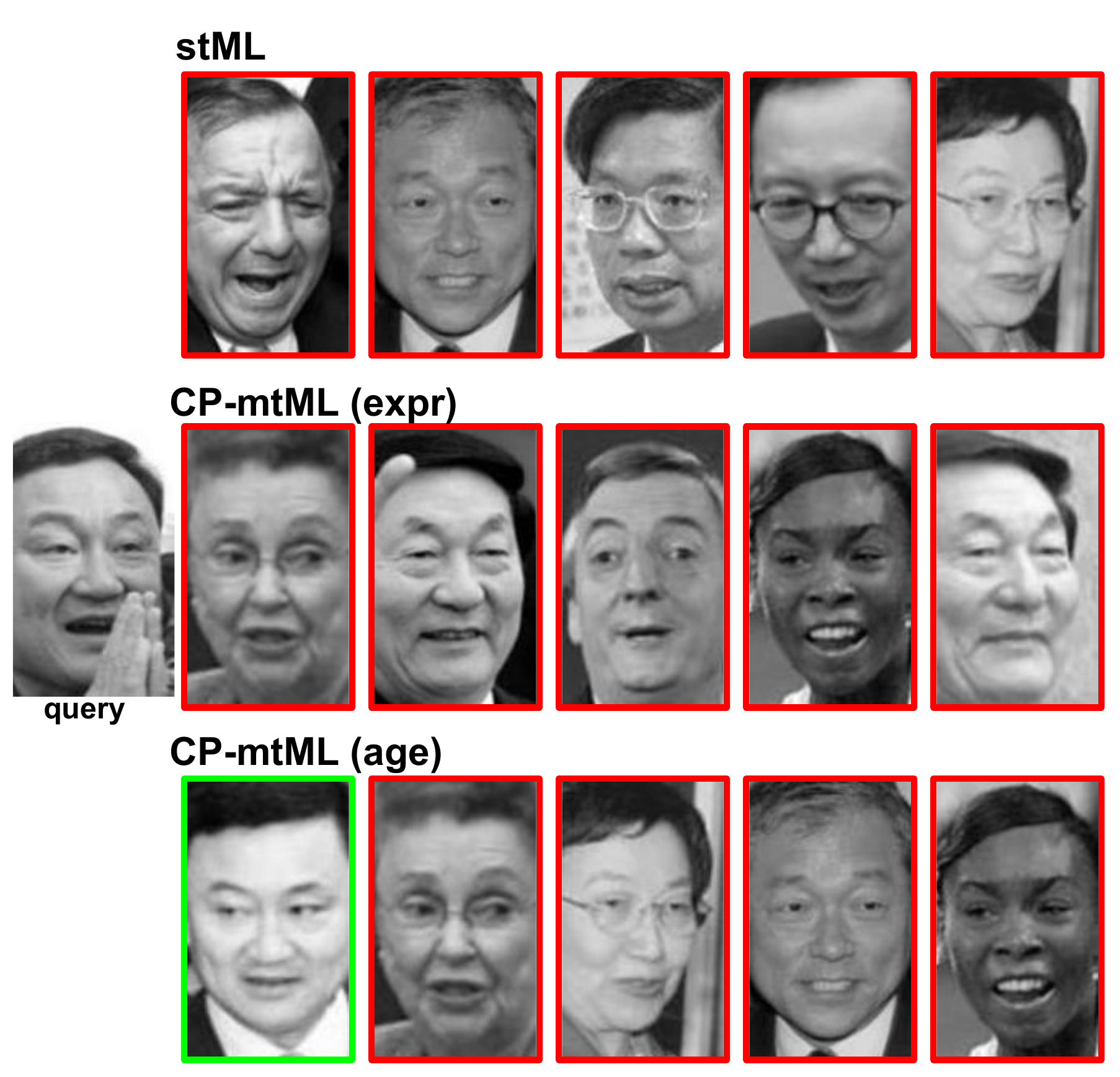} \hfill
\caption{
    \textbf{Sample set of queries for which CP-mtML (age) performs better than CP-mtML (expr) and stML.}
    The $5$ top scoring images (LBP \& no distractors) for the queries for the different
    methods. True (resp. false) positives are marked with a green (resp. red) border. Best viewed
    in color.
}
\label{fig1}
\end{figure*}

\begin{figure*}[t]
\centering
\includegraphics[width=0.32\textwidth, trim=0 40 200 0, clip]{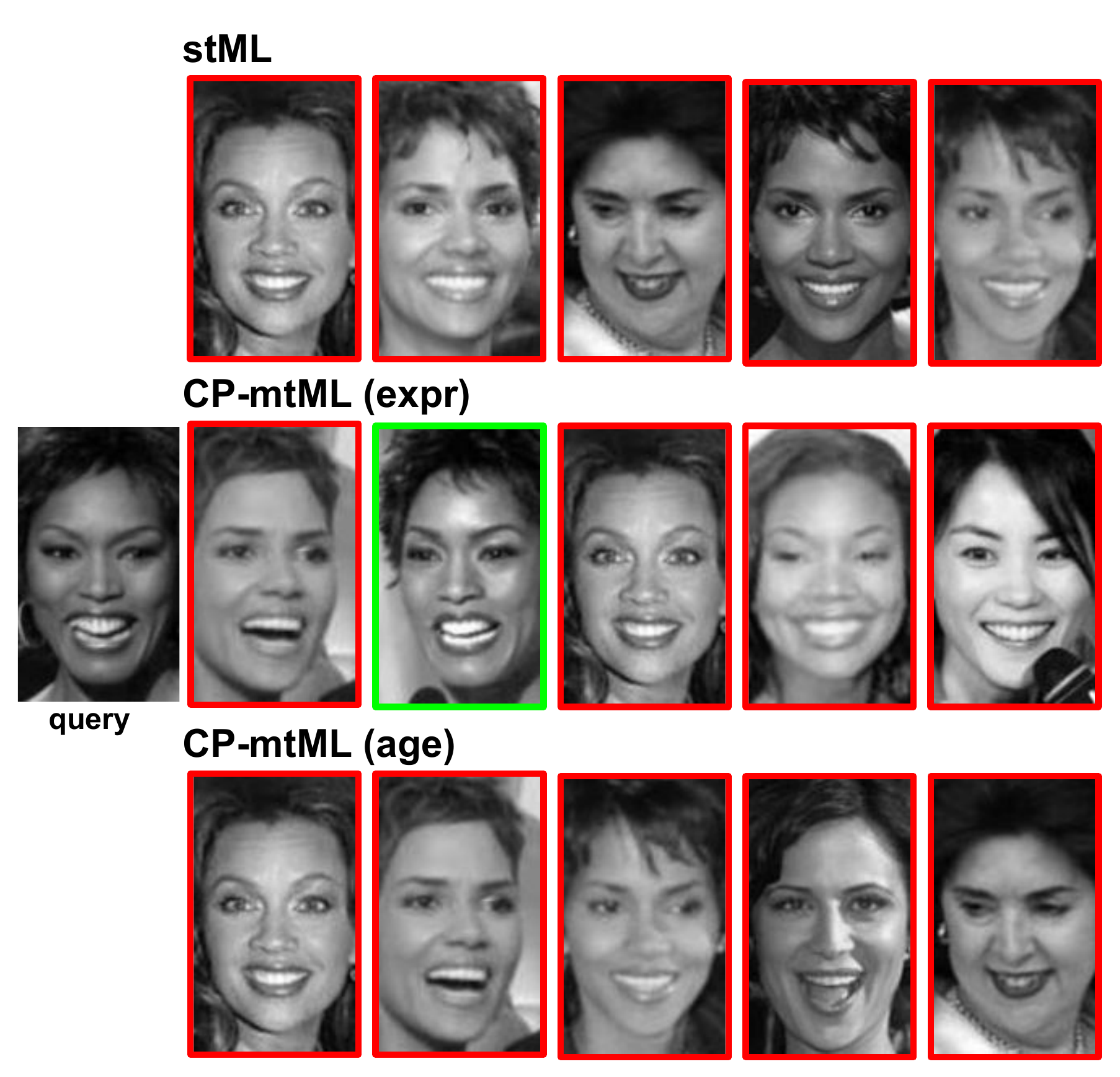} \hfill
\includegraphics[width=0.32\textwidth, trim=0 40 200 0, clip]{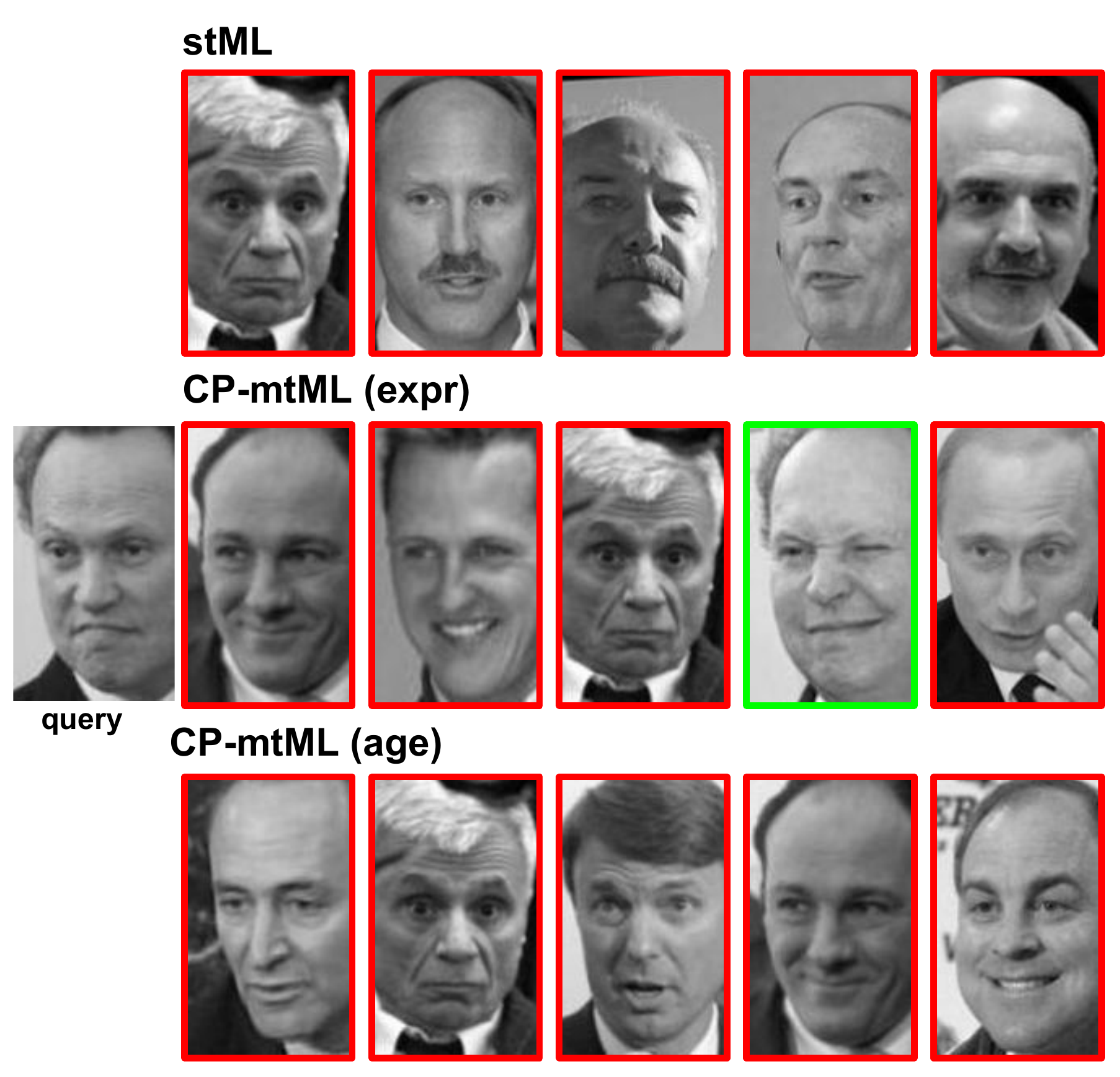} \hfill
\includegraphics[width=0.32\textwidth, trim=0 40 200 0, clip]{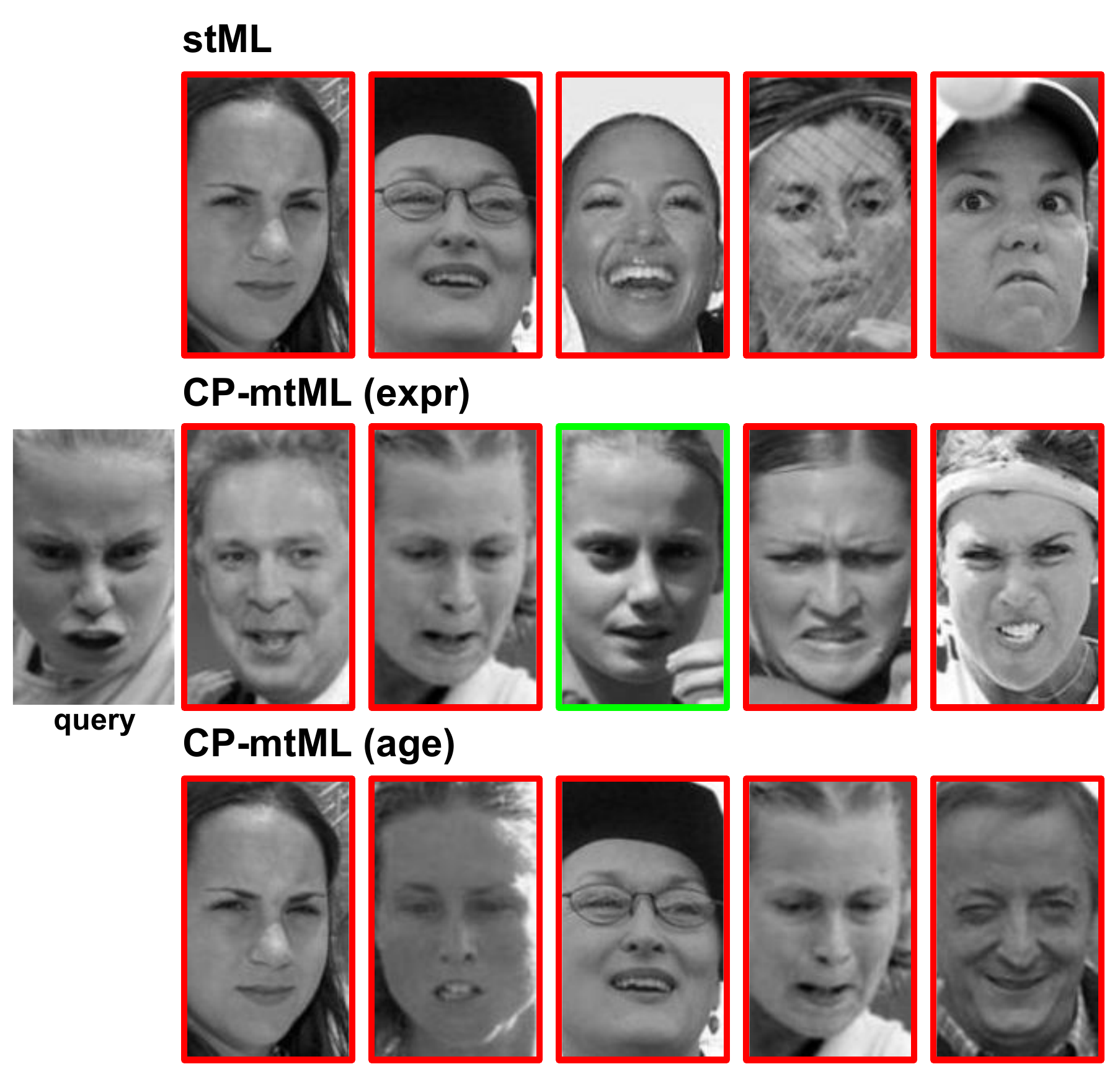}
\hfill
\caption{
    \textbf{Sample set of queries for which CP-mtML (expr) performs better than CP-mtML (age) and stML.}
    The $5$ top scoring images (LBP \& no distractors) for the queries for the different methods.
    True (resp. false) positives are marked with a green (resp. red) border. Best viewed in color.
}
\label{fig2}
\end{figure*}

\begin{figure*}[t]
\centering
\includegraphics[width=0.32\textwidth, trim=0 40 200 0, clip]{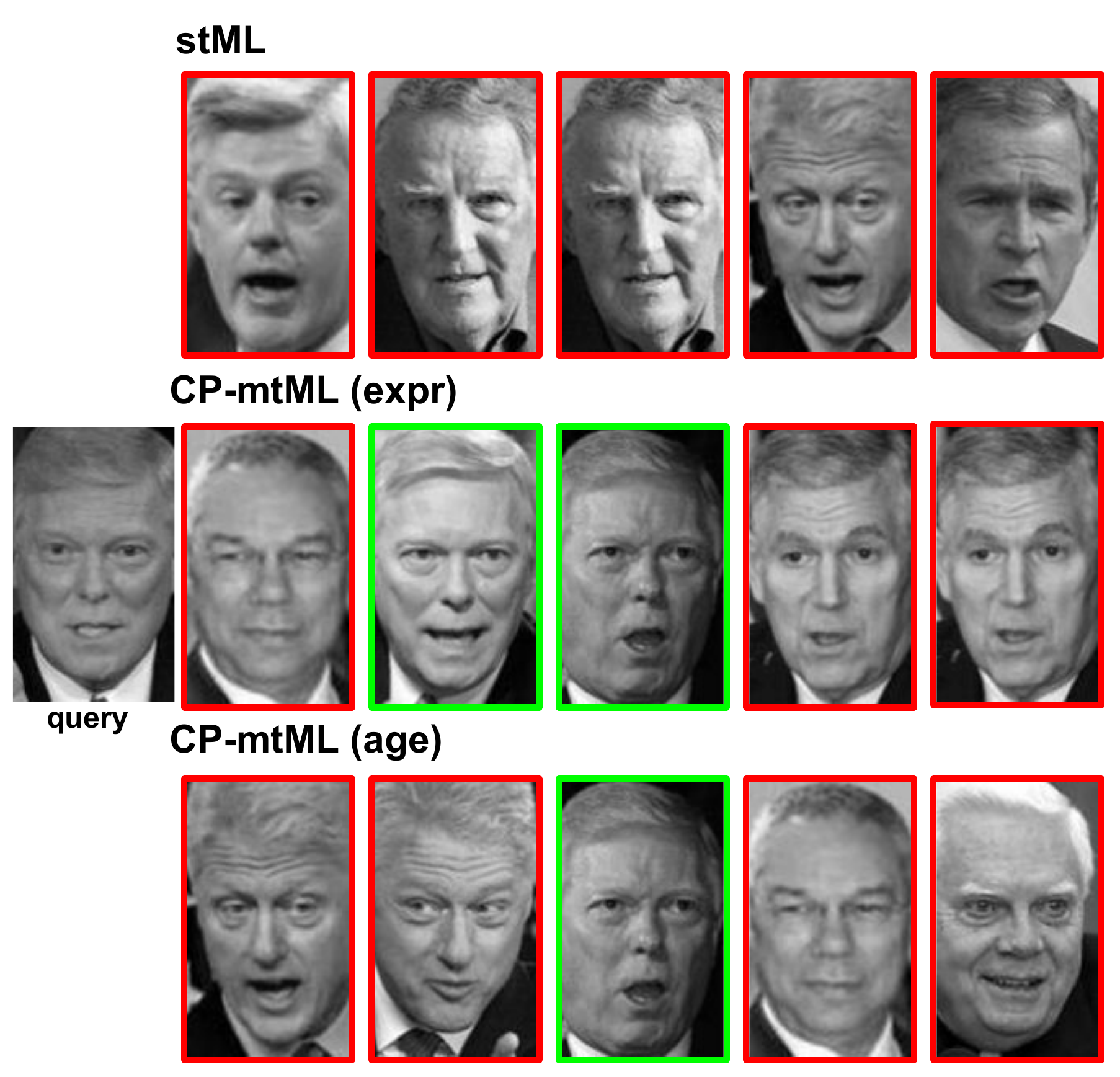} \hfill
\includegraphics[width=0.32\textwidth, trim=0 40 200 0, clip]{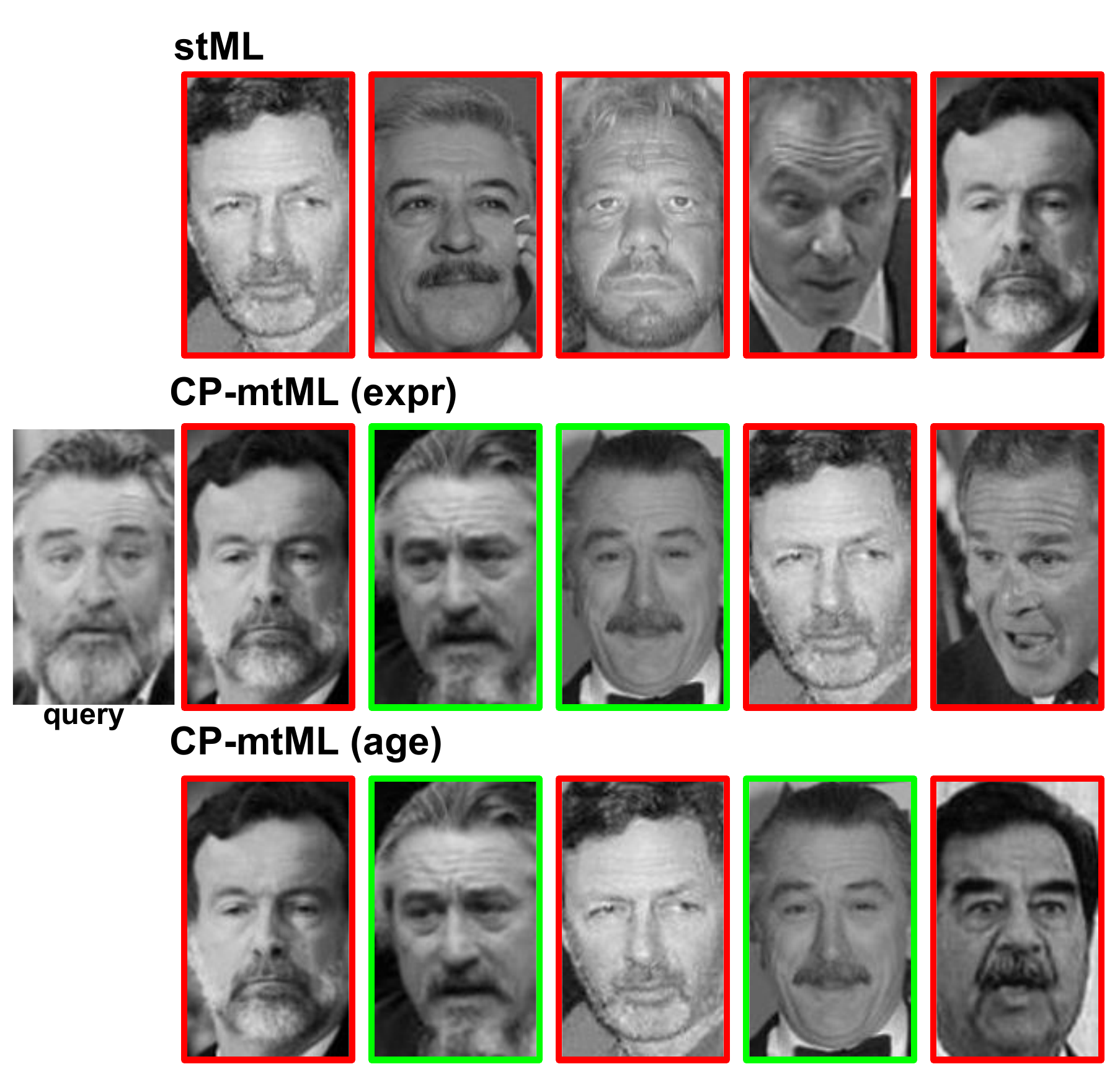} \hfill
\includegraphics[width=0.32\textwidth, trim=0 40 200 0, clip]{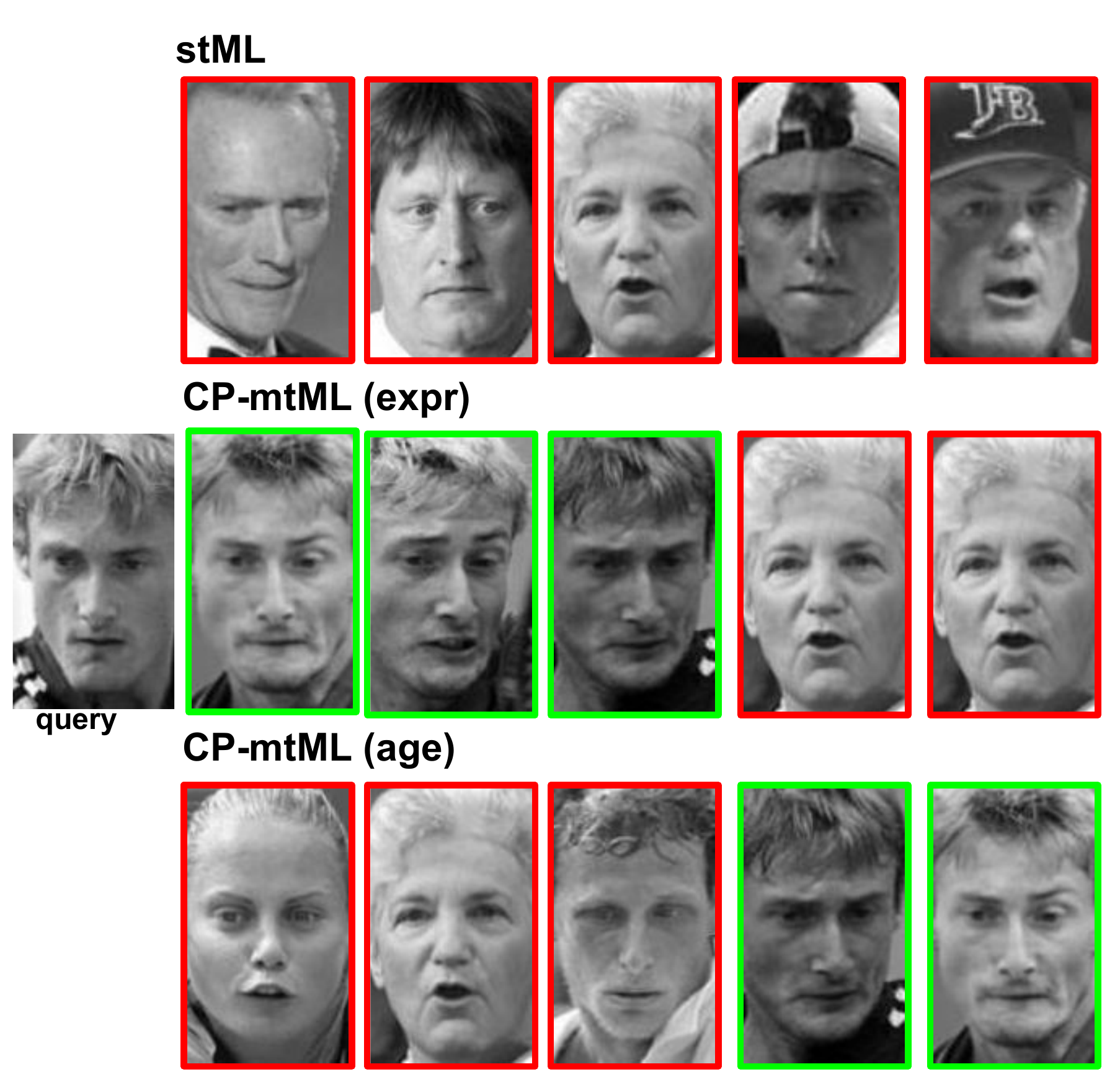}

\hfill
\caption{
    \textbf{Sample set of queries for which CP-mtML (expr) and CP-mtML (age) both perform better than stML. }
    The $5$ top scoring images (LBP \& no distractors) for the queries for the different methods.
    True (resp. false) positives are marked with a green (resp. red) border. Best viewed in color.
}
\label{fig3}
\end{figure*}

\begin{figure*}[t]
\centering
\includegraphics[width=0.32\textwidth, trim=0 40 200 0, clip]{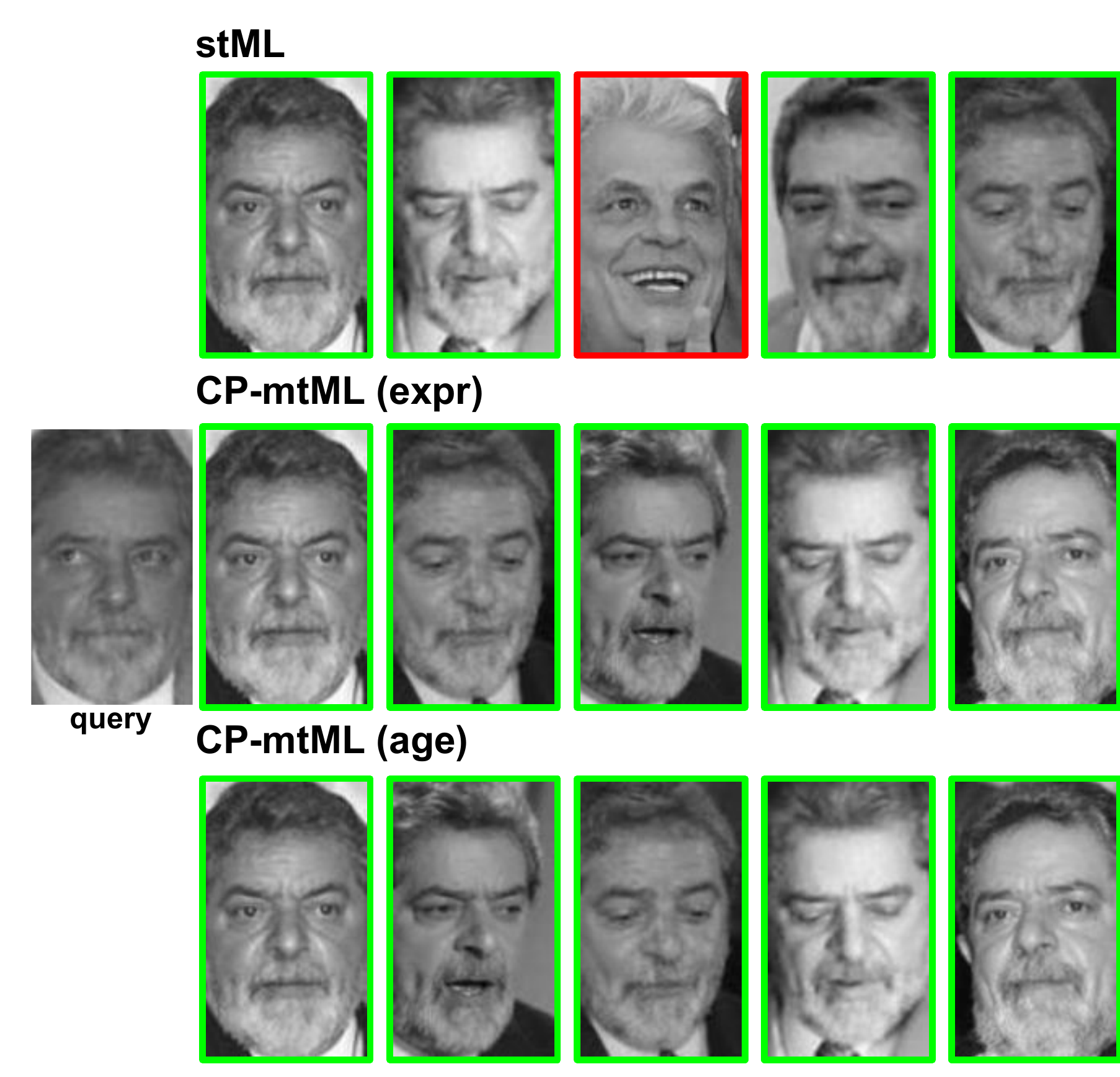} \hfill
\includegraphics[width=0.32\textwidth, trim=0 40 200 0, clip]{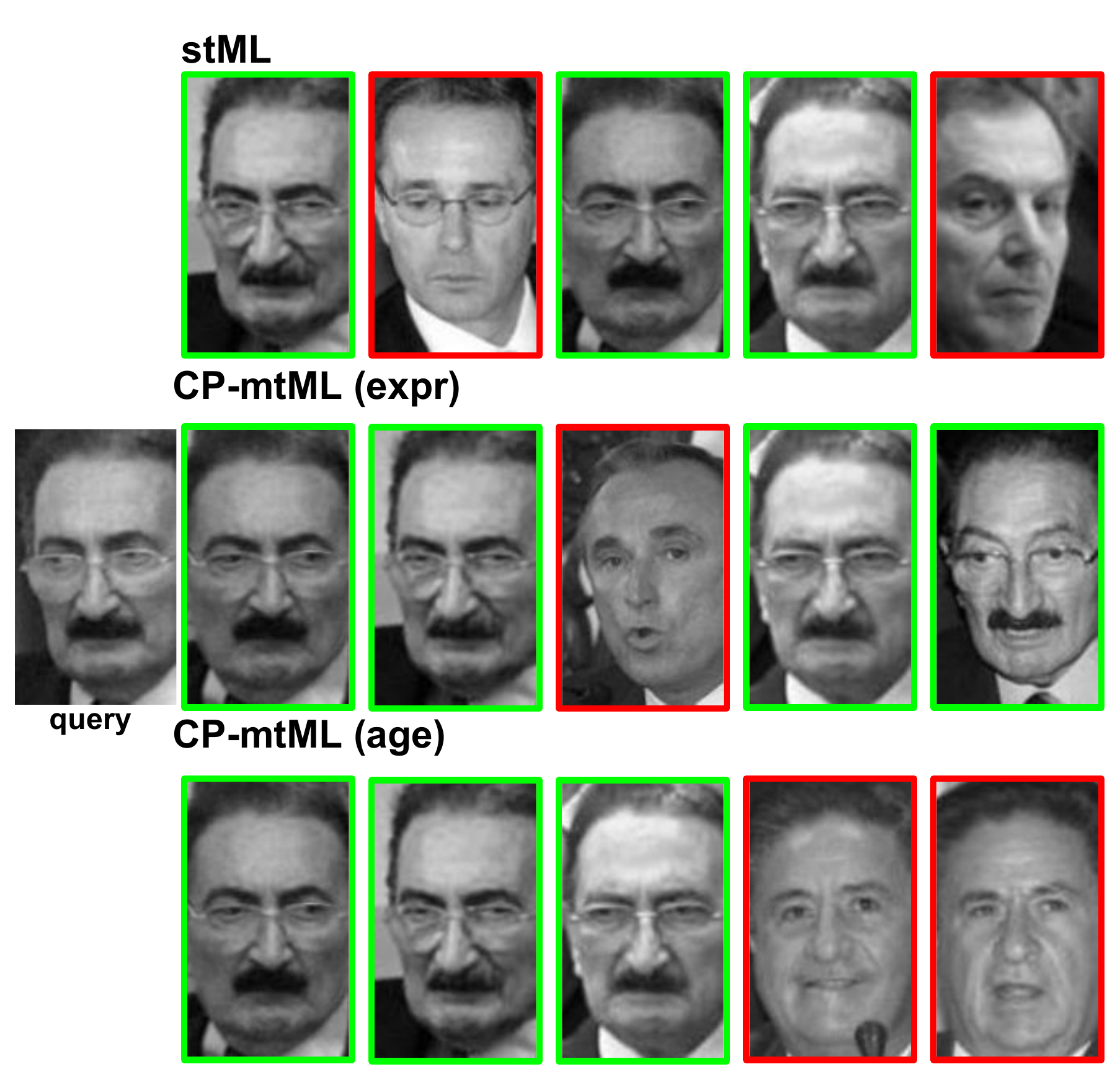} \hfill
\includegraphics[width=0.32\textwidth, trim=0 40 200 0, clip]{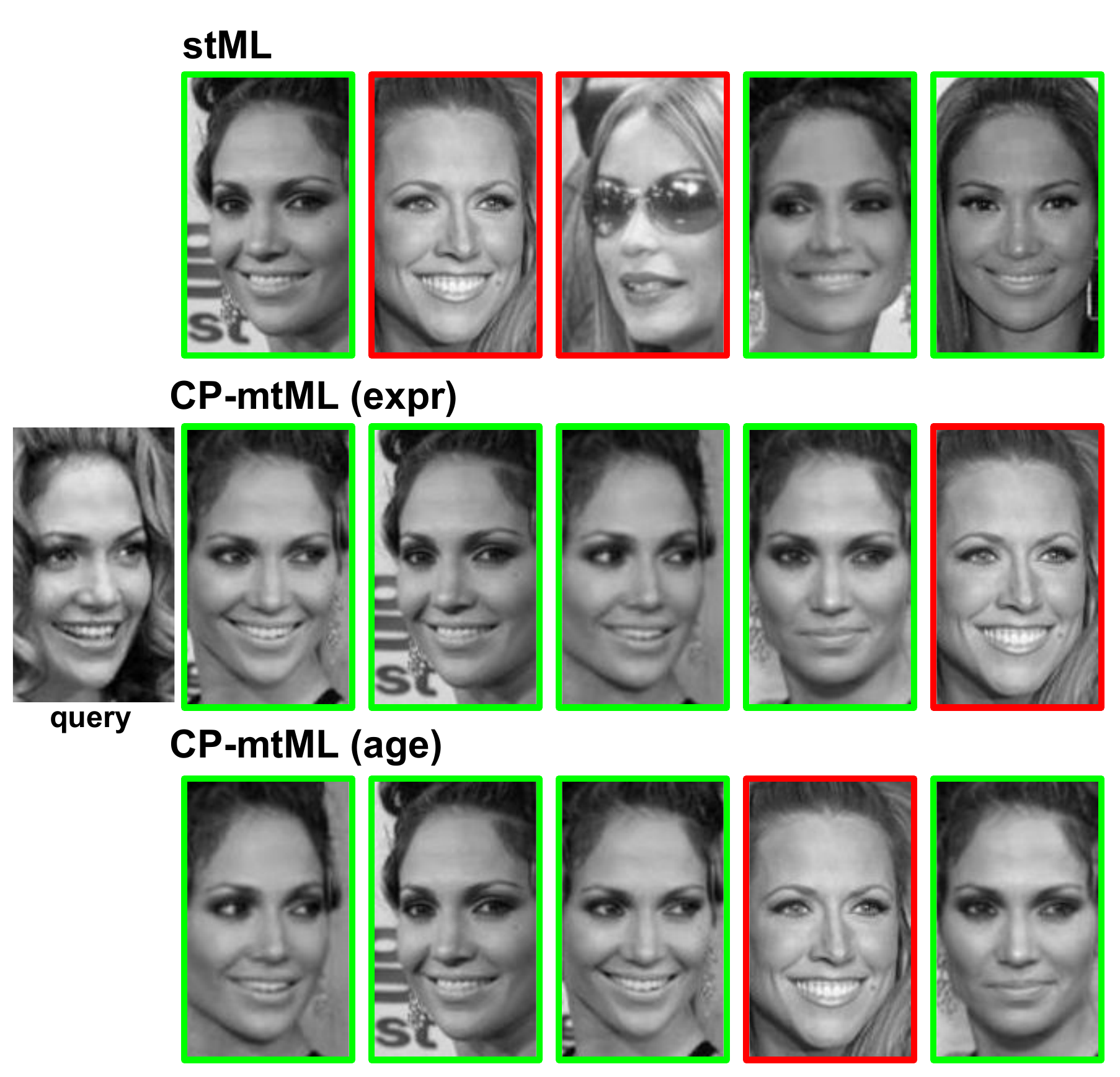}
\vspace{2em} \\
\includegraphics[width=0.32\textwidth, trim=0 40 200 0, clip]{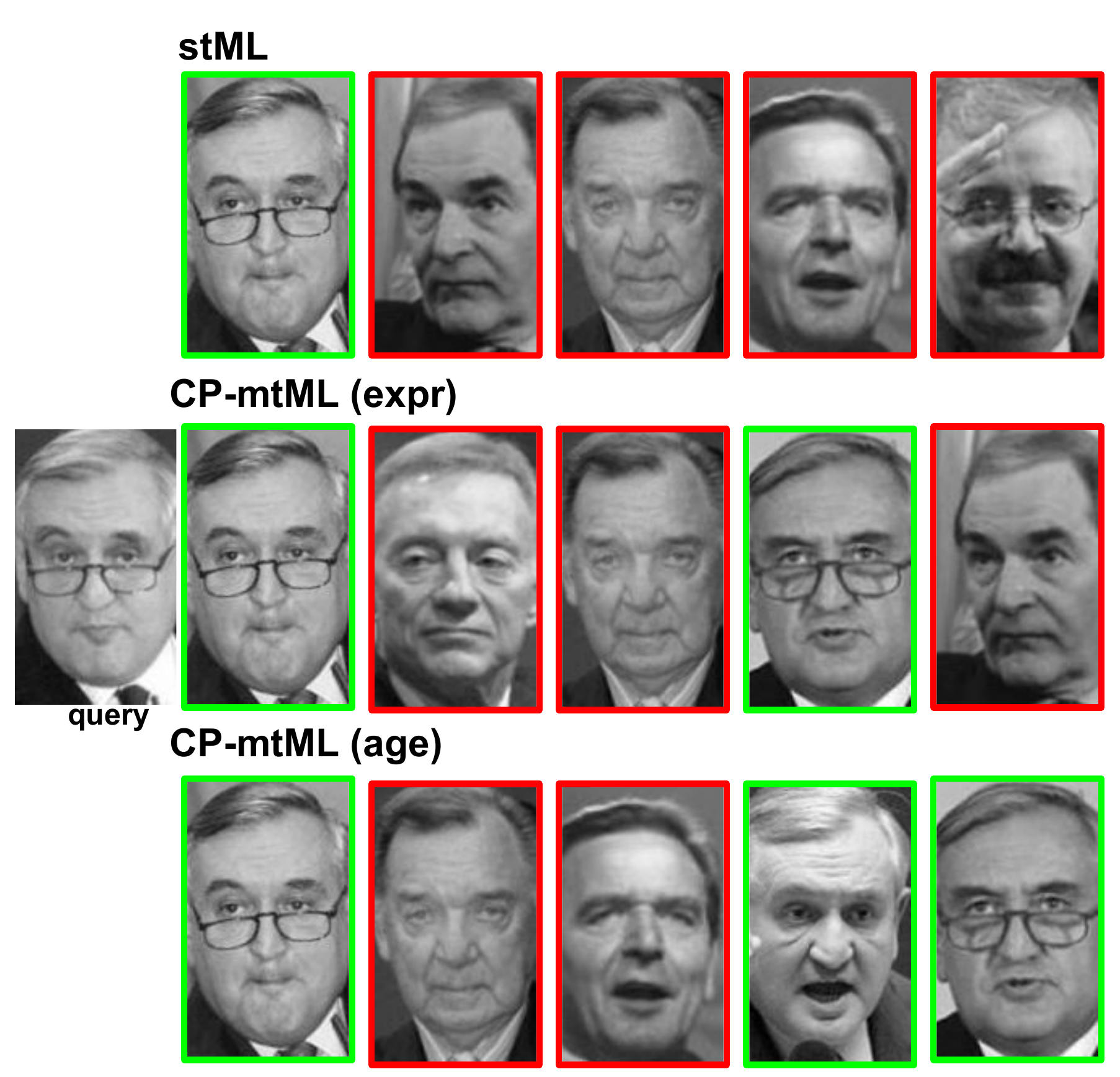} \hfill
\includegraphics[width=0.32\textwidth, trim=0 40 200 0, clip]{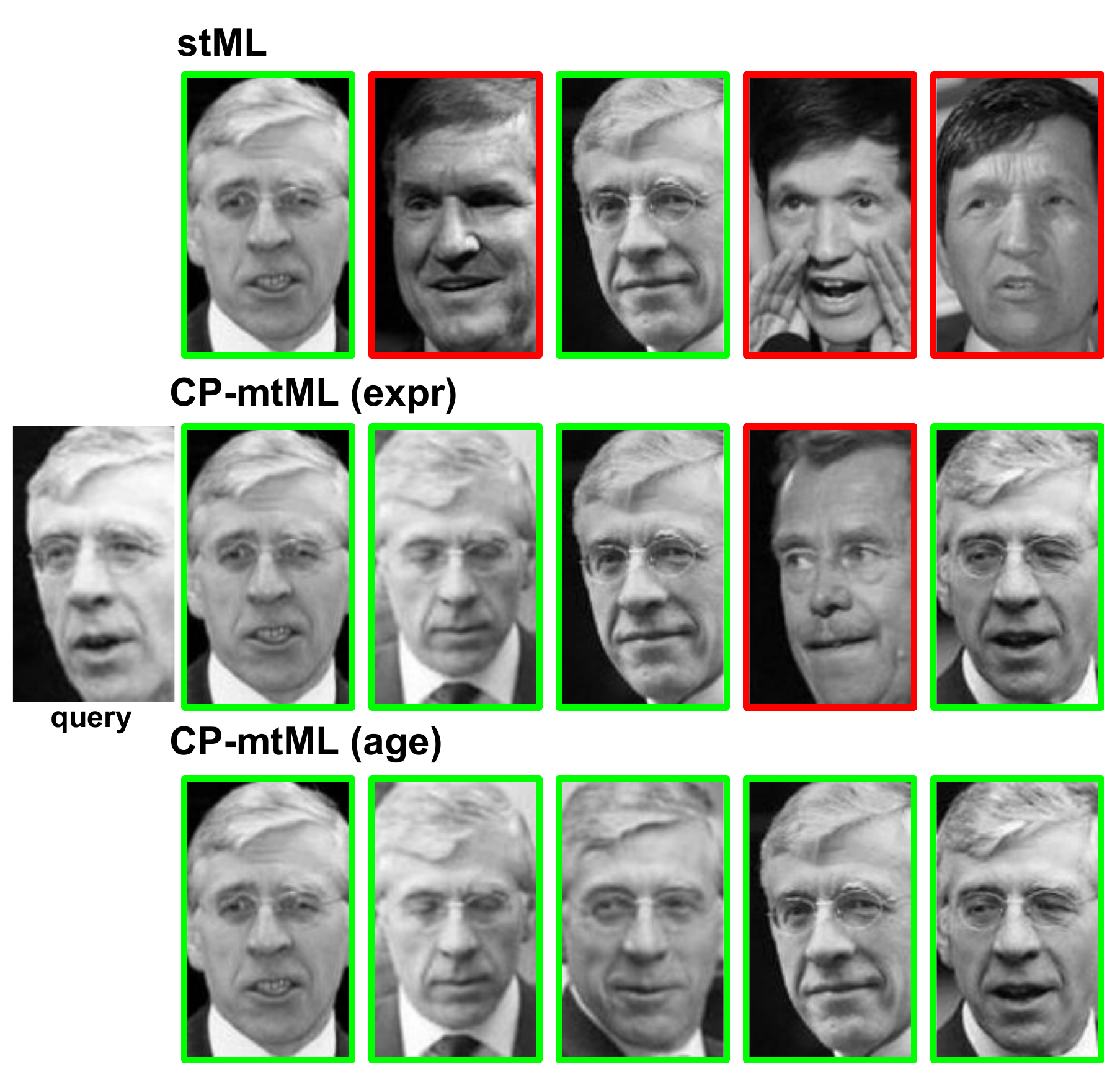}\hfill
\includegraphics[width=0.32\textwidth, trim=0 40 200 0, clip]{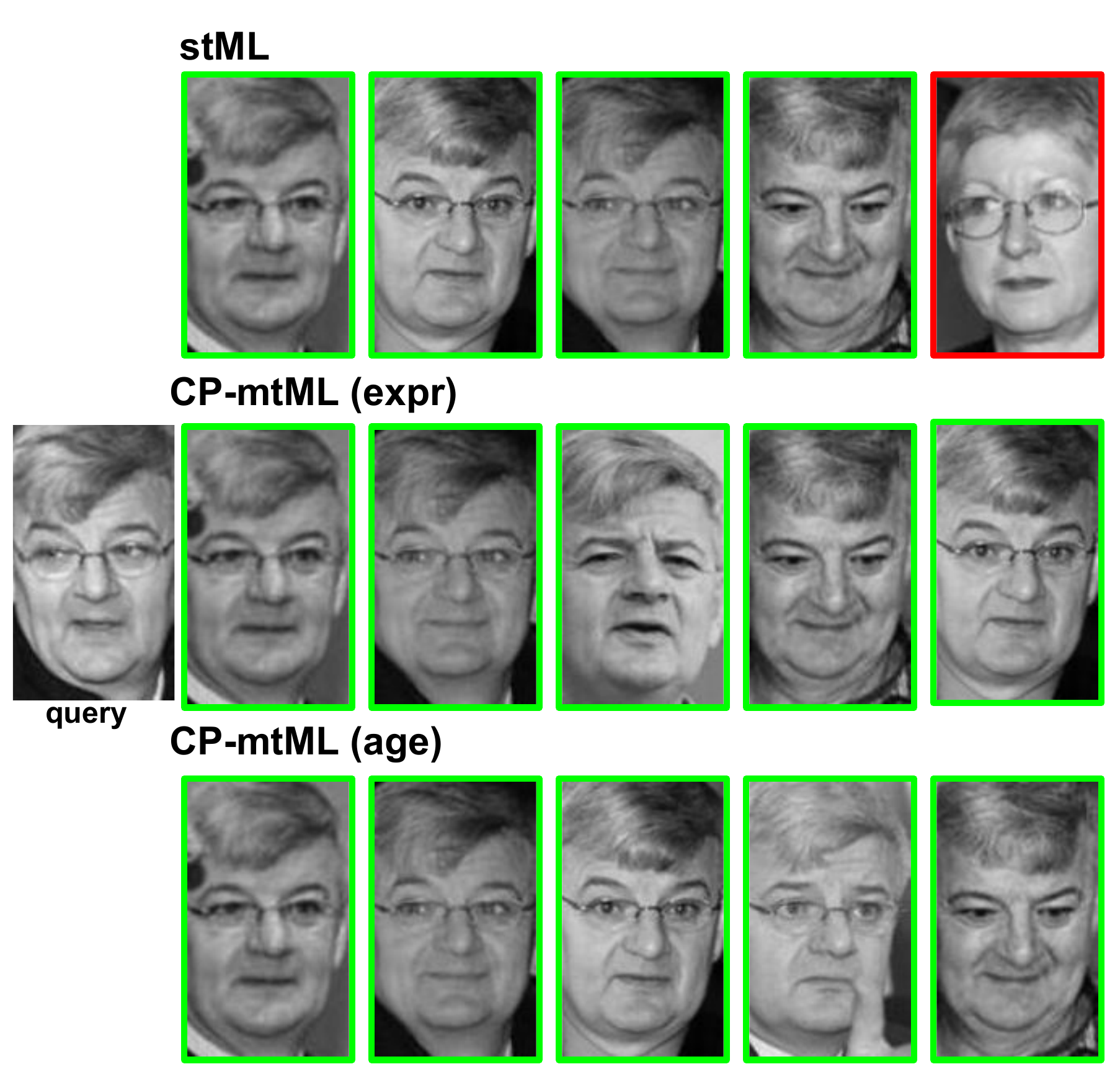}
\vspace{2em} \\
\includegraphics[width=0.32\textwidth, trim=0 40 200 0, clip]{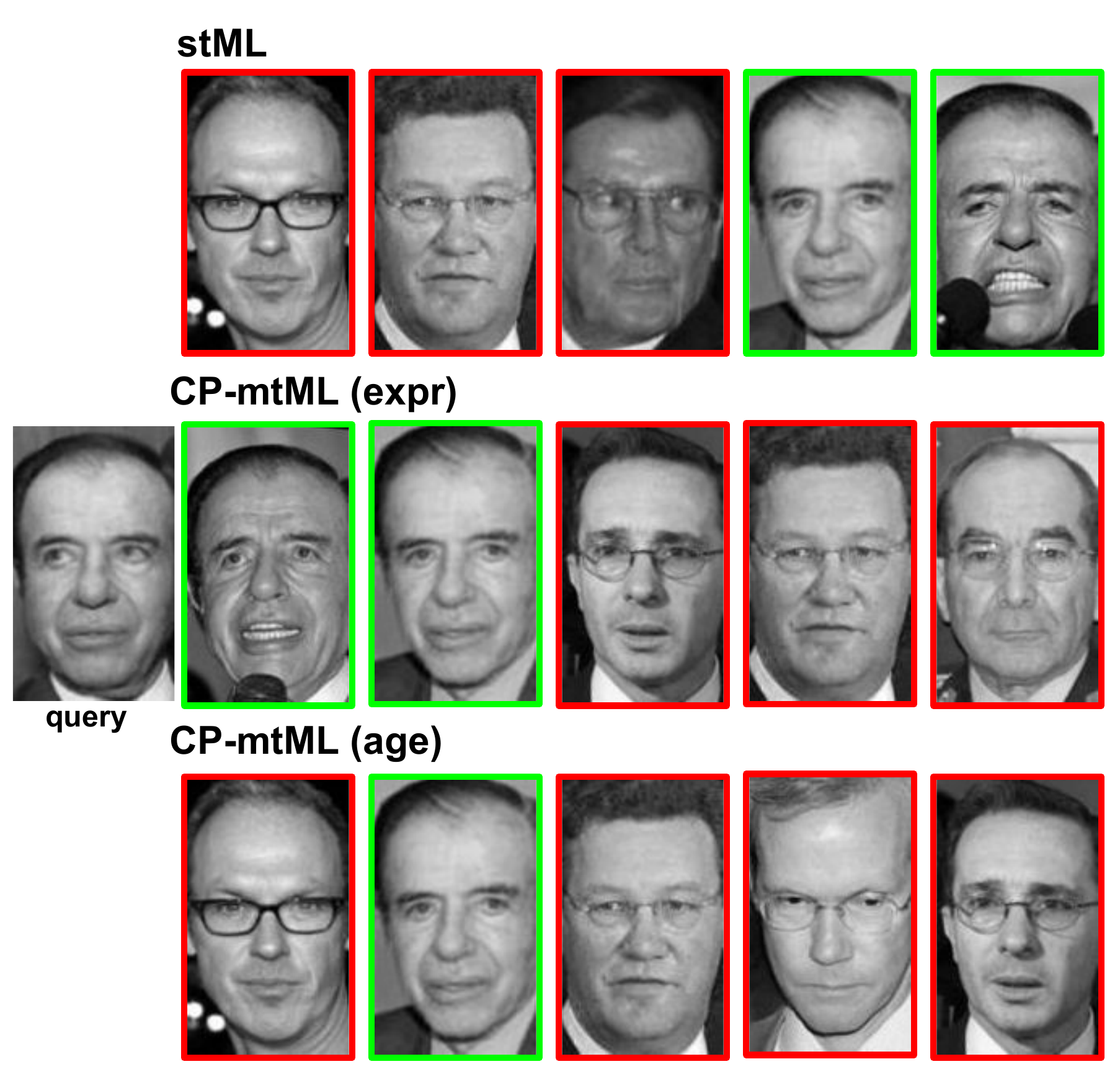} \hfill
\includegraphics[width=0.32\textwidth, trim=0 40 200 0, clip]{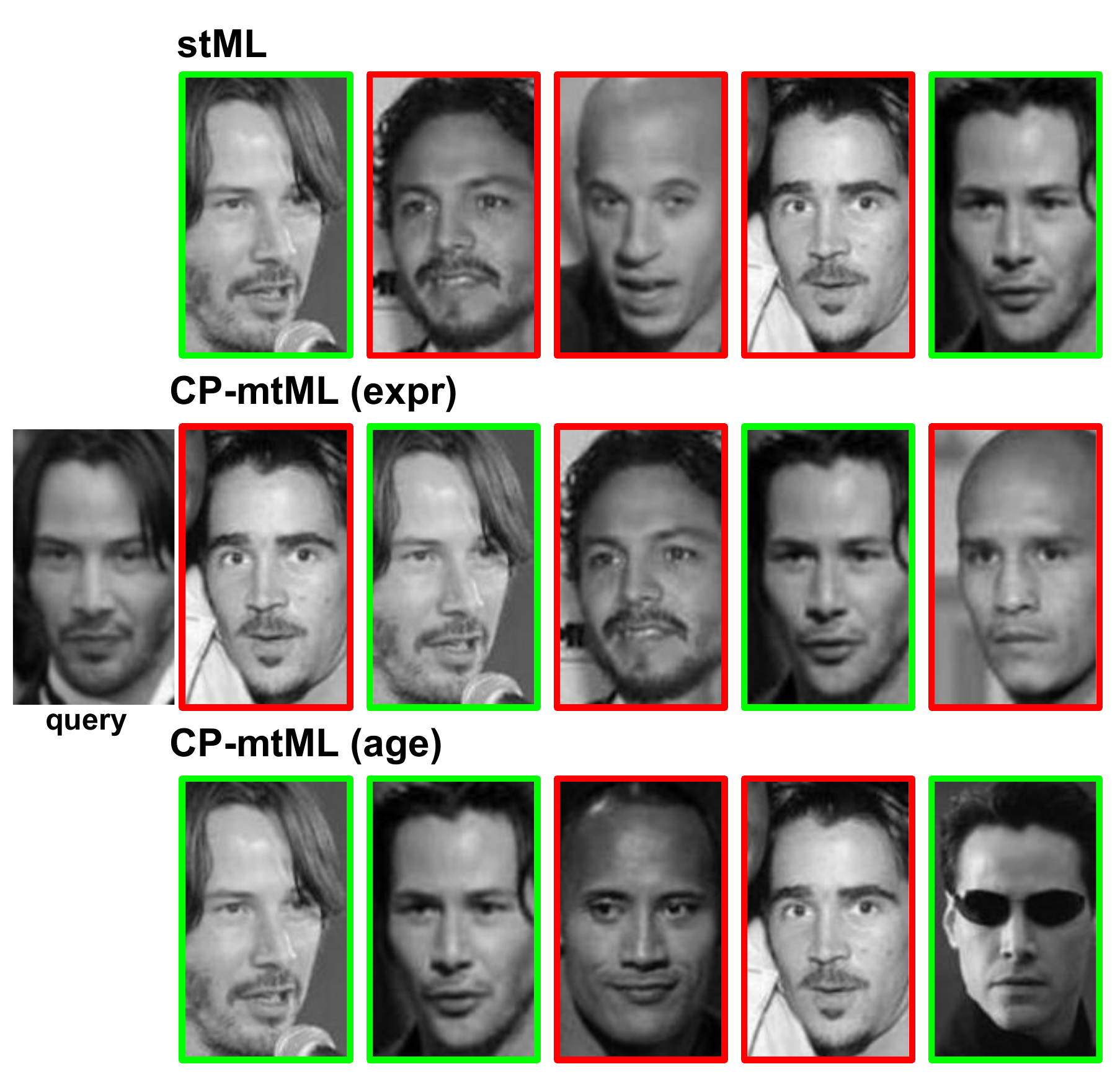}\hfill
\includegraphics[width=0.32\textwidth, trim=0 40 200 0, clip]{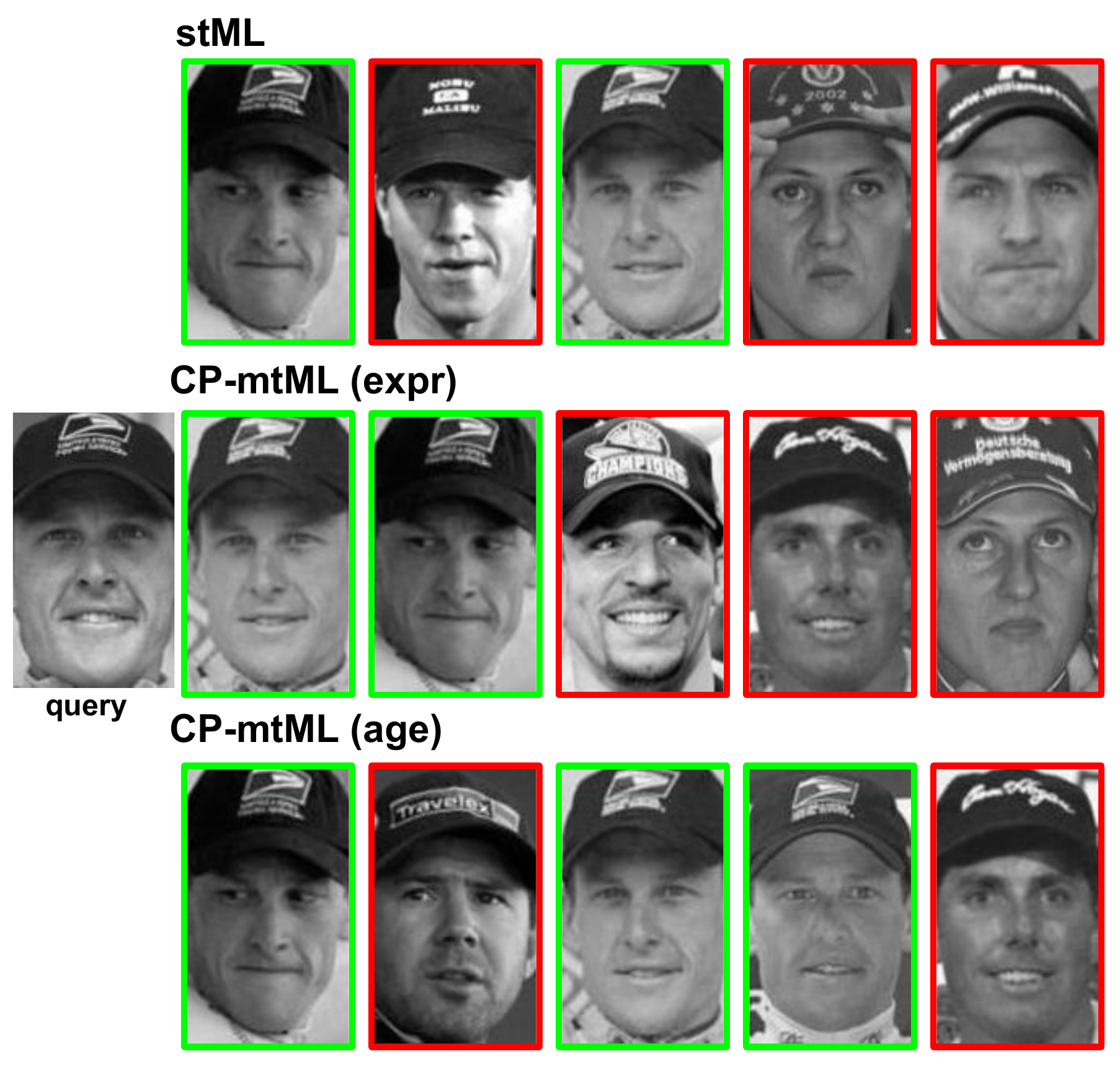}
\hfill
\caption{
    \textbf{Sample set of queries for which all of CP-mtML (expr), CP-mtML (age) and stML perform well.}
    The $5$ top scoring images (LBP \& no distractors) for the queries for the different methods.
    True (resp. false) positives are marked with a green (resp. red)
    border. Best viewed in color.
}
\label{fig4}
\end{figure*}

We can make the following observations 
\begin{enumerate}
\item Fig.~\ref{fig1} shows some queries for which CP-mtML (age) does better than CP-mtML (expr)
and stML. The results suggest that adding information based on age matching makes identity matching
more robust to high variations due to challenging pose (left) and occlusions (hair and hand in the
middle and right examples).

\item Fig.~\ref{fig2} shows some queries for which CP-mtML (expr) does better than CP-mtML (age)
and stML. The results suggest that adding information based on expression matching makes identity
matching more robust to challenging expressions.

\item Fig.~\ref{fig3} shows some queries for which CP-mtML (expr) and CP-mtML (age) do better than
stML. These cases are really challenging and the results retrieved by stML, while being sensible,
are incorrect. Adding more information based on age and/or expression matching improves results.

\item Fig.~\ref{fig4} shows some queries for which all three methods do well. These are queries
with either neutral expression and frontal pose or with characteristic appearances \eg moustache,
baseball cap, glasses, hairstyle \etc which occur for the same person in the gallery set as well.
\end{enumerate}

\end{document}